%% file: SPU-Net.tex
\documentclass[journal]{IEEEtran}

\ifCLASSINFOpdf
  \usepackage{microtype}
  \usepackage{graphicx}
  \usepackage{subfigure}
  \usepackage{booktabs} 
  \usepackage{amsmath}
  \usepackage{amssymb}
  \usepackage{bm}
  \usepackage{multirow}
  \usepackage{verbatim}
  \usepackage{makecell}
  \usepackage{authblk}
  \usepackage{bbding}
  \usepackage{color}
  \usepackage[colorlinks,linkcolor=red]{hyperref}
\else
\fi

\begin{document}
%
\title{Fast Learning Radiance Fields by Shooting Much Fewer Rays}

\author{Wenyuan Zhang,
        Ruofan Xing,
        Yunfan Zeng,
        Yu-Shen Liu,
        Kanle Shi,
        Zhizhong Han
\IEEEcompsocitemizethanks{
\IEEEcompsocthanksitem Wenyuan Zhang, Ruofan Xing are with the School of Software, Tsinghua University, Beijing, China. Yunfan Zeng is with the Department of Computer Science and Technology, Tsinghua University, Beijing, China. 
E-mail: zhangwen21@mails.tsinghua.edu.cn; xingrf20@mails.tsinghua.edu.cn; zengyf20@mails.tsinghua.edu.cn.
\IEEEcompsocthanksitem Yu-Shen Liu is with the School of Software, BNRist, Tsinghua University, Beijing, China. E-mail: liuyushen@tsinghua.edu.cn. Yu-Shen Liu is the corresponding author.
\IEEEcompsocthanksitem Kanle Shi is with  Kuaishou Technology, Beijing, China. E-mail: shikanle@kuaishou.com.
\IEEEcompsocthanksitem Zhizhong Han is with the Department of Computer Science, Wayne State University, USA. E-mail: h312h@wayne.edu.}
\thanks{This work was supported by National Key R\&D Program of China (2022YFC3800600, 2020YFF0304100), the National Natural Science Foundation of China (62272263, 62072268), and in part by Tsinghua-Kuaishou Institute of Future Media Data. Project Page is available at \url{https://zparquet.github.io/Fast-Learning} and code is available at \url{https://github.com/zParquet/Fast-Learning}.}}


\maketitle


\input{Abstract}
\input{Introduction}
\input{Related_Work}
\input{Method}
\input{Experiments}
\input{Conclusion}

\bibliographystyle{IEEEtran}
\bibliography{IEEEabrv,reference}

\end{document}

%% file: Abstract.tex
\begin{abstract}
Learning radiance fields has shown remarkable results for novel view synthesis. The learning procedure usually costs lots of time, which motivates the latest methods to speed up the learning procedure by learning without neural networks or using more efficient data structures. However, these specially designed approaches do not work for most of radiance fields based methods. To resolve this issue, we introduce a general strategy to speed up the learning procedure for almost all radiance fields based methods. Our key idea is to reduce the redundancy by shooting much fewer rays in the multi-view volume rendering procedure which is the base for almost all radiance fields based methods. We find that shooting rays at pixels with dramatic color change not only significantly reduces the training burden but also barely affects the accuracy of the learned radiance fields. In addition, we also adaptively subdivide each view into a quadtree according to the average rendering error in each node in the tree, which makes us dynamically shoot more rays in more complex regions with larger rendering error. We evaluate our method with different radiance fields based methods under the widely used benchmarks. Experimental results show that our method achieves comparable accuracy to the state-of-the-art with much faster training.
\end{abstract}


\IEEEpeerreviewmaketitle

%% file: Introduction.tex
\section{Introduction}

\begin{figure*}[htbp]
  \centering
  \includegraphics[width=\linewidth]{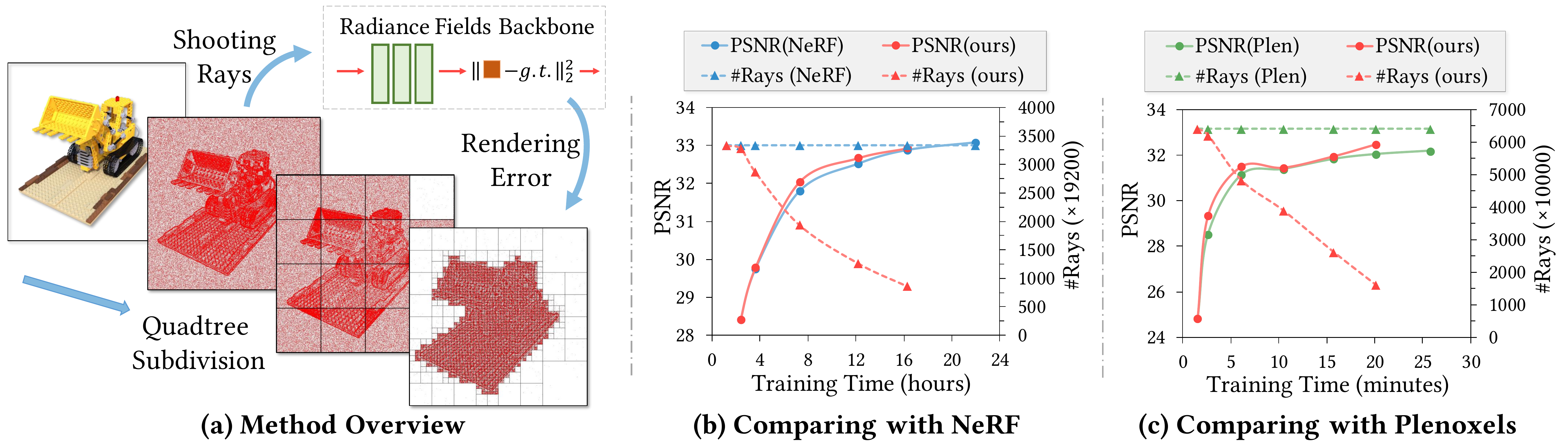}
  \caption{Illustration of our general strategy to learn radiance fields faster by shooting much fewer rays. \textbf{(a)} We use image context based probability distribution and adaptive quadtree subdivision to determine where to shoot more rays or to reduce rays on training images. Take ``lego'' in Synthetic dataset~\cite{mildenhall2020nerf} for example. \textbf{(b)} Based on NeRF~\cite{mildenhall2020nerf}, our method decreases the number of training rays from 3334 to 852, which reduces 30\% training time. \textbf{(c)} Based on Plenoxels~\cite{fridovich2022plenoxels}, our method decreases the number of rays from 12800 to 3218 and reduces 22\% training time. Both of our methods achieve a higher accuracy on test views.}
  \label{fig:teaser}
\end{figure*}

Learning a scene representation from single or multiple rendered views is an effective way to understand 3D shapes and scenes. Recent work has made great progress in reconstructing 3D meshes or point clouds from given 2D images for scene understanding~\cite{schonberger2016structure, liu2021fine, han2021hierarchical, wen20223d, handrwr2020, NeuralPull, Zhou2022CAP-UDF}. However, it is still a challenge to render realistic views from meshes and point clouds. As a solution, neural radiance field builds up a bridge between 2D images and 3D representations because it has the ability to both learn the implicit representation of 3D scenes and render novel views from any given perspectives
\cite{mildenhall2020nerf, zhang2020nerf++, yu2021pixelnerf, wang2021neus, xu2022point}.

A radiance field describes a scene by providing the color and density at each queried location in the scene. Hence, with a known camera pose, we can render a radiance field into an image from a specific view angle using volume rendering, which emits a ray on each pixel and accumulates the color and density of sampled points along each ray. Given multiple images taken from a scene as input, we can learn a radiance field of the scene by pushing the radiance field to be rendered as similar to the input images as possible. 

Current methods leveraging deep learning models to learn radiance fields have made significant progress. Radiance fields have been used in various applications such as novel view synthesis~\cite{mildenhall2020nerf, wang2021ibrnet, yu2021pixelnerf}, scene editing~\cite{yu2021unsupervised, yuan2022nerf}, stylization~\cite{zhang2022arf, huang2022stylizednerf} and generation~\cite{wang2022clip, jain2022zero, yao2022dfa}. Despite great success, the unaffordable learning time is still a big limitation suffering from the state-of-the-art methods.

Current methods speed up the learning of radiance fields by different strategies~\cite{SunSC22, fridovich2022plenoxels, mueller2022instant, Chen2022ECCV, hu2022efficientnerf}. For example, Plenoxels~\cite{fridovich2022plenoxels} replaced neural networks by a sparse voxel model to learn radiance fields, which achieves a speedup of two orders of magnitude compared to NeRF~\cite{mildenhall2020nerf}. TensoRF~\cite{Chen2022ECCV} models the radiance field of a scene as a 4D tensor, which represents a 3D voxel grid with per-voxel multi-channel features. Differently, Instant-NGP~\cite{mueller2022instant} introduced multiresolution hash encoding that permits the use of a smaller network without sacrificing quality, which also significantly reduces the training cost. Although these methods can train radiance fields fast, they require specific architectures, such as sparse voxel grids, discrete tensor coordinates or hash coding, which are not easy to adapt to improve the training efficiency of different neural radiance fields variations.

To resolve this issue, we introduce a general strategy to speed up the learning of radiance fields, as illustrated in Figure~\ref{fig:teaser}. Our key idea is to shoot much fewer rays to the radiance fields in volume rendering, which is a universal procedure during training in different neural radiance fields methods. Our contribution lies in our founding that we can perceive a radiance field without dramatically sacrificing accuracy by just shooting rays at pixels with dramatic color change, which significantly reduces redundancy of training rays from other pixels. Moreover, we also adaptively subdivide each view into a quadtree according to the average rendering error in each node in the tree, which makes us shoot more rays in more complex regions with larger rendering error. We evaluate our method with different radiance fields based methods under different benchmarks. Experimental results show that our method achieves comparable accuracy to the state-of-the-art with much faster training. Our contributions are listed below.

\begin{itemize}
  \item We present a method to speed up the learning of radiance fields. Our method can generally work for different radiance field based methods without requiring specific modifications.
  \item We justify the founding that the rays starting from different pixels have different perceiving ability of radiance fields.
  \item Our method can generalize to different radiance field based methods, and significantly reduce the training time under different scenes.
\end{itemize}

%% file: Related_Work.tex
\section{Related Work}
\noindent \textbf{Neural Radiance Fields.} \  Recently, NeRF~\cite{mildenhall2020nerf} has achieved impressive results in novel view synthesis task and has attracted lots of follow-up work. Unlike traditional explicit and discretized volumetric representations, NeRF uses a continuous 5D function to represent a static scene and optimizes a deep fully-connected neural network to learn this function. It can render high-resolution photorealistic novel views of objects and scenes from RGB images captured in natural settings. The follow-up work resolves the deficiencies, and expands possible applications of NeRF. NeRF++~\cite{zhang2020nerf++} extends NeRF to work for unbounded scenes, while NeRF-W~\cite{martin2021nerf} tackles the learning of radiance fields from unstructured photo collections. PixelNeRF, etc.~\cite{yu2021pixelnerf, jain2021putting, Niemeyer2021Regnerf, kim2022infonerf, rematas2021sharf} are devoted to solve the distortion problems when input views are sparse while others~\cite{su2021anerf, wang2021nerf, yen2021inerf} remove the requirement for pose estimation in volume rendering. In addition, there is still much work to study the solutions of defects and various applications of NeRF, e.g., mesh reconstruction~\cite{wang2021neus, oechsle2021unisurf, yariv2021volume}, relighting~\cite{zhang2021nerfactor, srinivasan2021nerv},
dynamic scene modeling~\cite{tretschk2021non, pumarola2021d, du2021neural, guo2021ad} and deformation~\cite{noguchi2021neural, park2021nerfies, park2021hypernerf}.

\noindent \textbf{Fast NeRF rendering.} \ Although NeRF has achieved amazing novel view synthesis effects, its training and rendering phase costs lots of time. Rendering a novel view usually takes about one minute, which is not conductive to real-time rendering and interaction. To solve this problem, NSVF~\cite{liu2020neural} uses both sparse voxel fields and classical techniques like empty space skipping and early ray termination to speed up the rendering procedure in NeRF. The idea of decomposition has also been adopted by many other methods~\cite{rebain2021derf, garbin2021fastnerf, reiser2021kilonerf}. DeRF~\cite{rebain2021derf} decomposes the scene into sixteen irregular Voronoi cells to speed up the rendering. FastNeRF~\cite{garbin2021fastnerf} splits the same task into two neural networks that are amenable to caching. Similarly, KiloNeRF~\cite{reiser2021kilonerf} decomposes the scene into thousands of tiny MLPs arranged on a regular 3D grid, leading to significantly higher speedups. Moreover, DONeRF~\cite{neff2021donerf} uses another network to predict the intersection of rays and mesh surfaces. Only 4 sampled points near surface are needed for neural rendering. AutoInt~\cite{lindell2021autoint} replaces the need for numerical integration in NeRF by learning a closed form solution of the antiderivative.

\noindent \textbf{Fast NeRF convergence.} \  
Training NeRF~\cite{mildenhall2020nerf} and its variants~\cite{zhang2020nerf++, martin2021nerf, wang2021nerf} usually takes 1 to 2 days, which limits its further applications. Many recent papers have proposed methods to improve the training efficiency of radiance fields. Some methods focus on rendering quality or sparse input views and bring faster convergence as a side benefit, such as generalizable pre-training~\cite{chen2021mvsnerf, wang2021ibrnet, kangle2021dsnerf}, anti-aliasing~\cite{barron2021mip, barron2022mip}, external depth~\cite{kangle2021dsnerf, liu2020neural} and so on. Other methods apply practical tricks to reduce training redundancy, such as skipping empty space~\cite{mueller2022instant, xu2022point, SunSC22}, pruning sample points on rays~\cite{mueller2022instant, hu2022efficientnerf, li2022nerfacc}, simplifying the two-stage coarse to fine training~\cite{barron2022mip, Chen2022ECCV, hu2022efficientnerf}, block decomposition~\cite{saragadam2022miner} and so on.

Recent studies~\cite{fridovich2022plenoxels, SunSC22, mueller2022instant, yu2021plenoctrees} adopt voxel grids to represent neural radiance field without a prior consultation. In these methods, parameters to be optimized are stored in voxel grid vertices. Parameters of sampled points are trilinear-interpolated from its neighboring vertices instead of directly using its 5D coordinates. Through this way, the amount of parameters to be optimized are reduced from infinite space to the same scale as the number of voxel grid vertices. Specifically, PlenOctrees~\cite{yu2021plenoctrees} extracts the NeRF structure into a sparse voxel grid in which each voxel represents view-dependent color using spherical harmonic (SH) coefficients. Plenoxels~\cite{fridovich2022plenoxels} extends PlenOctrees by additionally assign density parameters to voxel grids. The color and density of sample points are directly interpolated from grid vertices instead of predicted by MLP (Multi-Layer Perceptron). DVGO~\cite{SunSC22} achieves a similar speed as Plenoxels by splitting color and density into two voxel grids, where one is feature grid and the other is density grid.
The color is generated by a shallow MLP, using interpolated feature from feature grid as input, and the density is directly interpolated from density grid. 
Latest work Instant-NGP~\cite{mueller2022instant} uses hash encoding to store features in multiresolution voxel grids. The feature of a sample point is queried, interpolated and then concatenated from multiresolution features. It achieves both very fast speed and quite good rendering results in different tasks. 

Although the above methods have made some progress in terms of NeRF accelerating, they are task-specific and difficult to migrate to different radiance fields variations because of their specific and complex architectures. To resolve this issue, we introduce a general strategy to speed up the learning procedures for mainstream radiance fields based methods and achieves much faster training speed and comparable accuracy than the state-of-the-art works.

%% file: Method.tex
\section{Method}
\begin{figure*}[htbp]
  \centering
  \includegraphics[width=\linewidth]{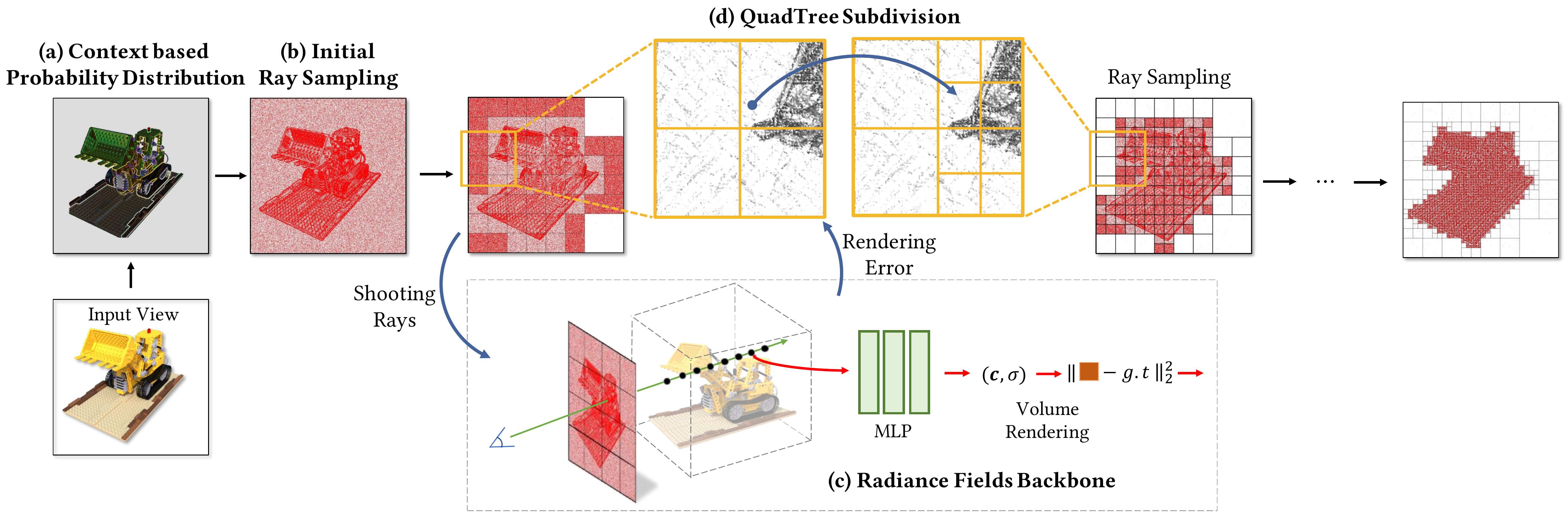}
  \caption{Overview of our method. Given an input view, we first (a) calculate a probability distribution according to the context around each pixel. Then we (b) use it as a prior probability distribution to sample pixels from which we shoot rays. These rays are fed into (c) a radiance fields backbone network and an average rendering loss is gathered for quadtree leaf nodes. Then (d) an adaptive quadtree subdivision algorithm is applied on each leaf node according to its average rendering loss to adjust the distribution of sampling rays.}
  \label{fig:network-structure}
\end{figure*}

Our method is a general framework of accelerating training procedure,
which can be easily integrated with mainstream 
radiance fields based methods. This generalization ability comes from the way of how we shoot much fewer rays in volume rendering to perceive radiance fields better, which is a common procedure in radiance fields based methods. Our method is formed by two main components: (1) a probability-based sampling function which samples rays according to the input image context and (2) an adaptive quadtree subdivision strategy that learns where to reduce rays in simple regions and where to increase rays in complex regions. Figure~\ref{fig:network-structure} illustrates our method, which will be detailed in the following.

\subsection{Volume Rendering}

NeRF~\cite{mildenhall2020nerf} proposed to use a continuous 5D function to represent a static scene and achieved great results in the view synthetic task.
We use the same differentiable model for volume rendering as in NeRF, where the color $\hat{\mathbf{c}}$ of a camera ray is rendered using 3D points sampled along the ray by the discretized volume rendering, defined by
\begin{equation}
  \begin{gathered} 
      \hat{\mathbf{c}} = \sum_{j=1}^N T_j(1-\exp(-\sigma_j \delta_j)) \mathbf{c}_j \\ 
      T_j = \exp(-\sum_{t=1}^{j-1} \sigma_t \delta_t),
  \end{gathered}
\end{equation}
%

\noindent where $\mathbf{c}_j$, $\sigma_j$ and $T_j$ represent the color, the opacity and the transmittance of the ray at $j$-th sampled point, respectively, $\delta_j$ indicates the distance between $j$-th and $(j+1)$-th adjacent sampled points. This function for calculating $\hat{\mathbf{c}}$ from the set of $(\mathbf{c}_j, \sigma_j)$ values is trivially differentiable and reduces to traditional alpha compositing with alpha values $\alpha_j=1-\exp(-\sigma_j \delta_j)$.

\subsection{Ray Sampling Probability}
Specifically, given a 3D point $\mathbf{x}=(x,y,z)\in \mathbb{R}^3 $ and a related viewing direction $\mathbf{d}=(\theta, \phi)\in \mathbb{R}^2$, a neural network $F_\Theta: (\mathbf{x}, \mathbf{d})\mapsto (\mathbf{c}, \sigma)$ maps a 5D point $(\mathbf{x}, \mathbf{d})$ to an emitted color $\mathbf{c}=(r,g,b)$ and a volume density $\sigma$. The parameters of the network are optimized by minimizing the pixel-level color error obtained by volume rendering.

During training, an amount of 5D points are sampled on each ray $\mathbf{r}_i^k\in \mathbf{R}_i$, where $\mathbf{r}_i^k$ represents the $k$-th ray from the $i$-th training image $I_i$, and the set $\mathbf{R}_i$ of the rays is generated from $I_i \in \mathcal{I}, i=1,2...N$, where $N$ is the number of images and $\mathcal{I}$ is the set of all the training images. Volume rendering $V_\Theta: \mathbf{r}_i^k\mapsto \hat{\mathbf{c}}_i^k$ is then used for rendering per-pixel color $\hat{\mathbf{c}}_i^k$ using color and density predicted from the sampled 5D points along $\mathbf{r}_i^k$. The object function is to minimize the following volume rendering loss from all the emitted rays
\begin{equation}
    \mathcal{L}=\frac{1}{NM}\sum_{I_i\in \mathcal{I}} \sum_{\mathbf{r}_i^k\in \mathbf{R}_i} ||V_\Theta (\mathbf{r}_i^k)-\mathbf{c}_i^k||_2^2 ,
\end{equation}

\noindent where $\mathbf{c}_i^k$ is the pixel's ground truth color and $M$ is the size of ${\mathbf{R}_i}$, that is, the amount of emitted rays in each image $I_i$.

With regard to how to obtain $R_i$ of all the training images, almost all radiance fields based methods randomly select $M$ pixels in the image $I_i$, and then emit $M$ rays from the selected pixels along the viewing direction. 
Given a pixel $(u, v)$, we use $\mathbf{r}_i(u, v)$ to represent a ray starting from the pixel and the ray orientation is consistent with the viewing direction of image $I_i$. In this way, the sampled ray positions obey uniform distribution on the training images:
\begin{equation}
    \mathbf{r}_i(u, v) \sim {U(I_i)}, u\in [0, H_i], v\in [0, W_i],
\end{equation}

\noindent where $H_i$ and $W_i$ are the height and width of image $I_i$, respectively.

Although such distribution is straight and proved to be effective in practice~\cite{mildenhall2020nerf, zhang2020nerf++}, there are still some potential problems. (1) The uniformly distributed rays cannot well capture the 
non-uniformly distributed information. In practice, there are many areas of continuous pixels with the similar color on images, which we call \emph{trivial areas}, such as the background of synthetic images, the sky in outdoor scene images, the wall in indoor scene images. Pixels in such kinds of trivial areas usually contain less information due to semantic consistency with their neighborhood pixels. 

As a result, the rendering error of the rays shot from trivial areas can converge fast, because the density of most sampled points along the rays is close to 0. Therefore, we only need to shoot fewer rays in the trivial areas where the color changes slightly to perceive the radiance fields. (2) In contrast, pixels in the \emph{nontrivial areas} where color changes greatly contain more information, so more rays are required to capture the detailed information and learn how to distinguish these pixels' colors from its neighboring ones.

\begin{figure*}[htbp]
  \centering
  \includegraphics[width=\linewidth]{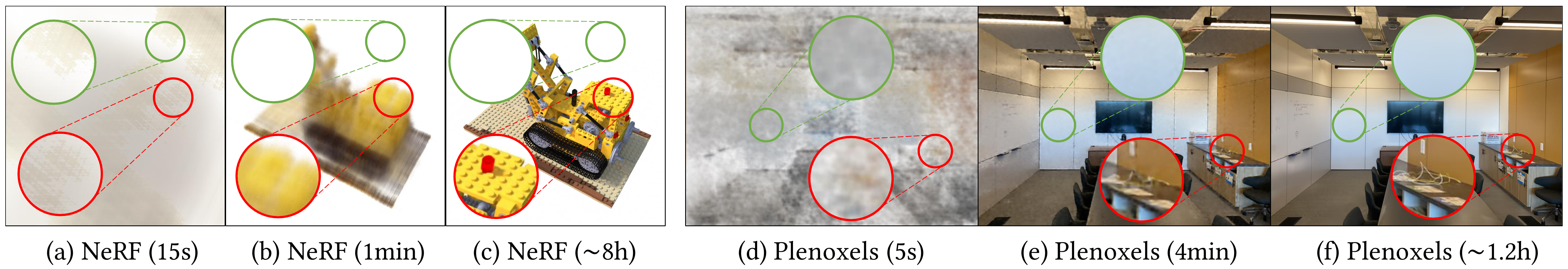}
  \caption{Comparison of convergence speed in different areas. (a) - (c) is the training process of NeRF~\cite{mildenhall2020nerf} on the synthetic dataset. The blank background area converges rapidly in 1 minute, while the lego building block area doesn't become clear until about 8 hours of training. (d) - (f) is the training process of Plenoxels~\cite{fridovich2022plenoxels} on the forward-facing dataset. Similarly, the wall of the room becomes the same color in 4 minutes, while the wires on the table cannot be distinguished until 1.2 hours.}
  \label{fig-method3.2}
\end{figure*}

Figure~\ref{fig-method3.2} illustrates an example. We observe that the learning of radiance fields in trivial areas (highlighted by green circles) converges in a very short time, while it takes a long time to finely render the nontrivial areas (highlighted by red circles) with a great change in colors. Therefore, there is no need to keep shooting a large number of rays at the trivial areas for training.
In contrast, it makes more sense to shoot more rays in nontrivial areas to perceive the radiance field. Based on the above observation, we propose two strategies to optimize ray distribution on input images. The first one is to calculate a prior probability distribution based on the image context, and the second one is to apply an adaptive quadtree subdivision algorithm to dynamically adjust ray distribution.


\subsection{Context based Probability Distribution \label{sec: colorProb}}

In order to identify the trivial and nontrivial areas in an image, we introduce the \emph{image context} to capture the area importance of this image, 
which will be used to non-uniformly sample rays on the image.
Specially, we use a center surround mechanism for computing image context, because it has the intuitive appeal of being able to identify areas
that are different from their surrounding context. 
Intuitively, a pixel, that has the same color with its neighbors, has only a few context. In contrast, a pixel, whose color is quite different from its surrounding pixels, has more context. 
Therefore, we use the color variation of pixels relative to their surroundings to quantitatively identify the image context.

Given an input view before training, we design a probability density function $g(u, v)$, which maps a pixel $(u, v)$ position into a prior probability, indicating how likely a ray is sampled at this position. We define $g$ as the standard variation of the color $\mathbf{c}$ of pixel $(u, v)$ and its 8 one-order neighboring pixels,
\begin{equation}
\label{eq:std}
    \begin{gathered} 
        g(u, v)={\rm std}(\mathbf{c}(u, v))=\sqrt{\frac{1}{9} \sum_{x, y}[\mathbf{c}(x, y)-\Bar{\mathbf{c}}]^2}, \\
        x\in \{u-1, u, u+1\}, \ y\in \{v-1, v, v+1\}.
    \end{gathered}
\end{equation}
where $\Bar{\mathbf{c}}$ is the mean color among the center pixel $(u, v)$ and its 8 adjacent pixels. 


In 2D image space, pixels with high values of $g$ usually correspond to the areas where color changes greatly. Accordingly, in 3D space, these pixels usually correspond to the positions where density changes greatly, and the positions are often on the boundary surfaces of 3D objects. How to accurately detect the surface of objects has proved to be significant to 3D reconstruction and novel views synthesis~\cite{wang2021neus, oechsle2021unisurf, yariv2021volume}, and our image context based probability distribution function naturally helps to estimate where surfaces are located.

To balance the difference between maximum and minimum values of $g$, we utilize a normalization operation on $g$ below
\begin{equation}
    g'(u, v)=\frac{\operatorname{clamp}(s, \max(g(u, v)))}{\max(g(u, v))},
    \label{eq: probability-normalization}
\end{equation}

\noindent where we typically define threshold $s=0.01\times \operatorname{mean}(g(u, v))$. 
Values less than $s$ will be clamped to $s$ to avoid sampling too few rays at the corresponding positions. 
After normalization, $g'(u, v)$ is distributed in interval $[0, 1]$.  We use the distribution $g'$ instead of uniform distribution to generate rays for neural radiance field training, i.e.
\begin{equation}
    \mathbf{r}_i(u, v) \sim g'(u, v), u\in [0, H_i], v\in [0, W_i].
\end{equation}
where $\mathbf{r}_i(u, v)$ is the ray emitted from pixel $(u, v)$ on $i$-th image.

Compared with the uniform distribution, our probability distribution takes into account the context information of the image. We sample more rays in areas with large probability and sample less rays in the areas with small probability. Figures~\ref{fig:quadtree-syn}, \ref{fig:quadtree-llff}, \ref{fig:quadtree-tnt} provide several visualization examples of our sampling strategy. In the lines of ``Sampled Rays Distribution'', we sample 50\% rays according to the context based probability distribution and randomly sample the other 50\% rays, where each red point represents a sampled ray. It can be observed that more rays are distributed in the areas with complex shapes and colors, i.e. nontrivial areas, where color changes a lot. On the other hand, fewer rays are distributed in trivial areas, such as the white background, the walls and the floors.

\subsection{Adaptive QuadTree Subdivision \label{sec: quadtree}}

\begin{figure}[htbp]
  \centering
  \includegraphics[width=\linewidth]{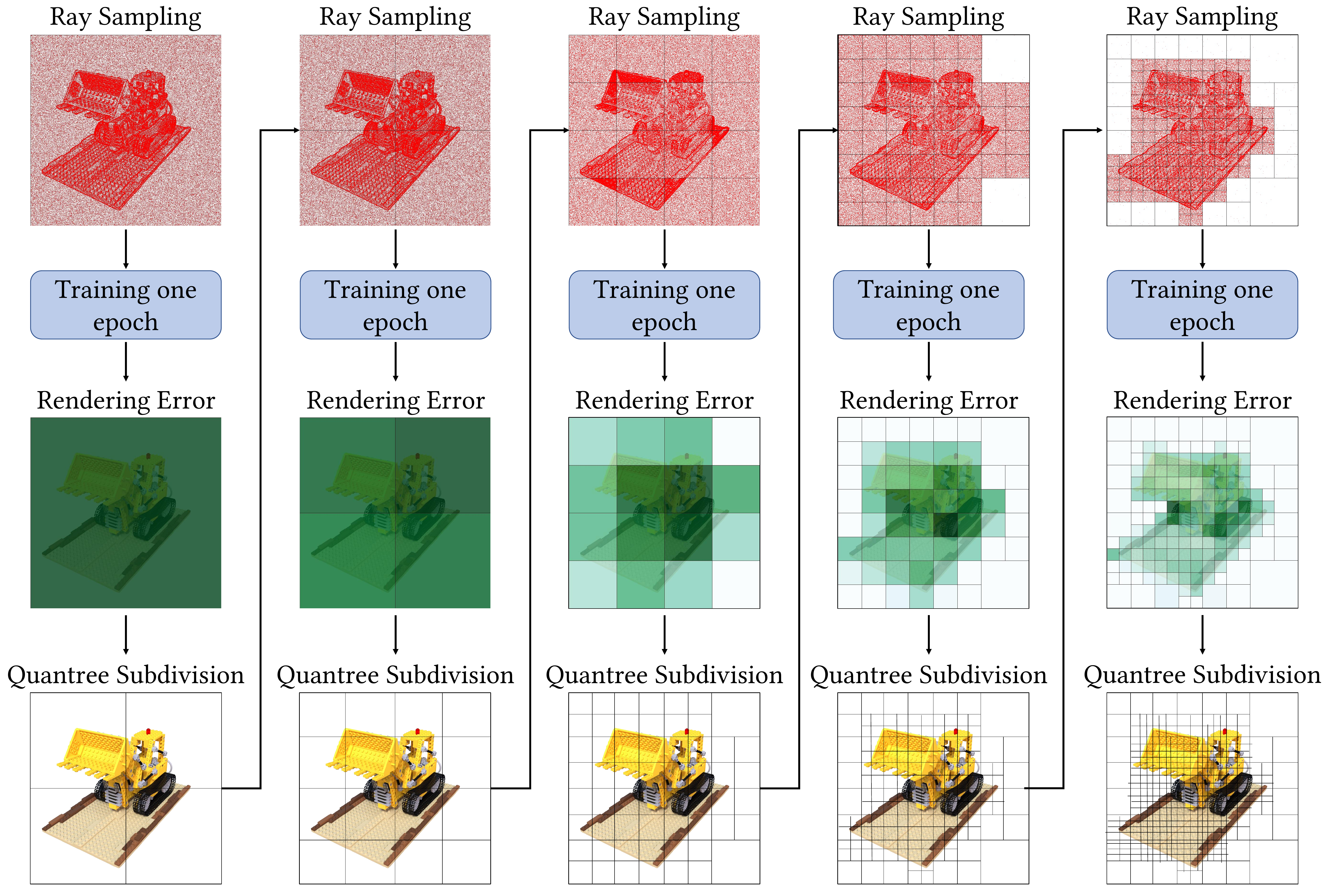}
  \caption{A demonstration of the overall procedure of ray sampling, training, obtaining rendering error and quadtree subdivision.}
  \label{fig:quadtree-subdivision}
\end{figure}

Based on the prior probability distribution on images, we further utilize an adaptive quadtree subdivision to adjust the position of shooting rays and to reduce the number of training rays. We firstly construct a quadtree on each input view, in which each leaf node represents one block of the image. As the training procedure goes on, we constantly traverse every leaf node. For each node $F$, we first sample and shoot rays on $F$ according to the prior probability distribution, and then calculate the average rendering error $e_F$ over all the emitted rays $\mathbf{r}_i\in F$ by
\begin{equation}
    e_F=\operatorname{mean}_{\mathbf{r}_i\in F}(V_\Theta(\mathbf{r}_i)-\mathbf{c}_i) ,
\end{equation}
where $\mathbf{c}_i$ is the ground truth color of the pixel emitting $\mathbf{r}_i$. If $e_F$ is smaller than the pre-defined threshold $a$, we conclude that the training in this block has converged well, so the leaf node $F$ is marked to be left aside. Only a few rays will be sampled on $F$ and it won't be subdivided again anymore. Otherwise, if the average rendering error $e_F$ is larger than threshold $a$, we conclude the rendered color is still far from the ground truth pixel color, which indicates that further training is still needed, so this node will be subdivided into four smaller leaf nodes. The selection of threshold $a$ will be discussed in detail in ablation study in Section~\ref{sec:ablation}.

Let $M_i$ denote the number of rays emitted from image $I_i$. Before subdivision, rays are randomly selected on image $I_i$ at the times of total number of pixels, so $M_i=H_i\times W_i$ at the beginning of training, where $H_i$ and $W_i$ are the height and width of $I_i$. Let $Q_1^l$ and $Q_2^l$ denote the number of unmarked leaf nodes and marked leaf nodes separately, and $l$ denote the times of subdivision.
In an unmarked node, the number of sampled rays is equal to the number of pixels of the node, while in a marked node, only a few rays are sampled. Then we get the total number of emitted rays after $l$ times of quadtree subdivision:
\begin{equation}
    M_i^l = Q_1^l\times \frac{H_i}{2^l}\times \frac{W_i}{2^l} + Q_2^l \times n_0 ,
\end{equation}

\noindent where $n_0$ is a constant number of rays sampled on marked leaf nodes (we typically set $n_0=10$ in practice). With the increase of subdivision times, the growth rate of $Q_2^l$ is far lower than that of $Q_1^l$. As a result, much fewer rays are sampled on marked leaf nodes, so the total amount of rays to be trained will gradually decrease. We draw a flow chart in Figure~\ref{fig:quadtree-subdivision} to illustrate the overall process of ray sampling, training, rendering error and quadtree subdivision as below. This figure will help clarify the sampling and quadtree subdivision procedure.

We noticed that iMap~\cite{sucar2021imap} similarly uses rendering error as sampling guidance. Here we clarify the main differences between iMap and our sampling strategies. (1) \emph{The applications of rendering loss are different.} iMap uses the rendering loss distribution on image blocks to decide how many points should be sampled on each block, while we use the rendering loss on each leaf node to decide whether this node should be subdivided into 4 child nodes. (2) \emph{The number of sampled points and image blocks are different.} In our method, the number of sampled points in each leaf node is identical, but the number and area of the image blocks (i.e. leaf nodes) changes during training. In contrast, iMap samples different numbers of rays according to a render loss based distribution in each one of the same size blocks. (3) \emph{The sampling strategies in image blocks are different.} In each image block, iMap uniformly samples points for rendering, while we sample points according to the image context. More points are sampled in the nontrivial areas where color changes a lot, while fewer points are sampled in the trivial areas where color changes slightly. Our sampling strategy helps to capture the detailed information in the nontrivial areas and reduce the training burden in the trivial areas. The ablation study (Table~\ref{tab:ablation-quadtree}: “Ours w/o prob”) further demonstrates that our context-based sampling method is efficient and effective. In conclusion, although rendering loss is both used in our method and iMap, there are essential differences between these two methods on the applications of rendering loss, the number and sampling strategies of points, and the problems to be solved.

\begin{figure*}[htbp]
  \centering
  \includegraphics[width=\linewidth]{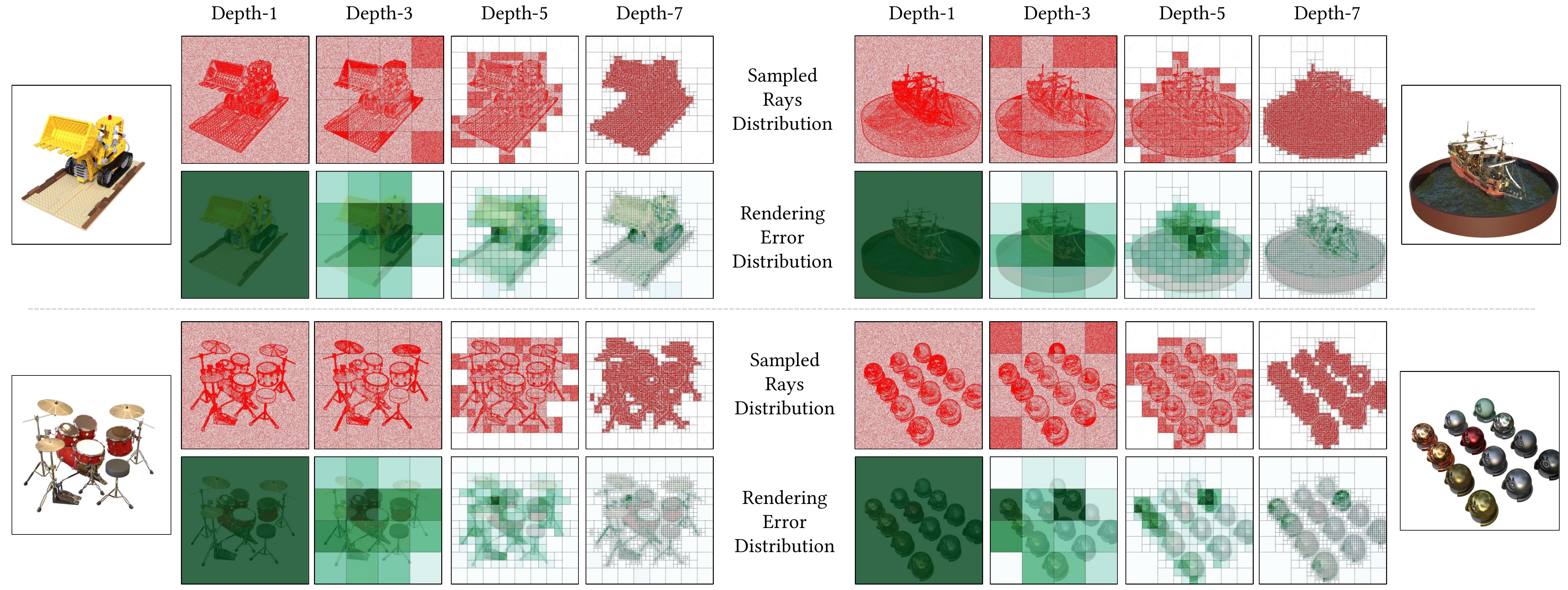}
  \caption{Visualization of sampled rays distribution and rendering error distribution at different quadtree depths. We select ``Lego'', ``Ship'', ``Drums'', ``Materials'' in Synthetic Dataset~\cite{mildenhall2020nerf} as examples. In the lines of ``Sampled Rays Distribution'', each red point represents a sampled ray and in the lines of ``Rendering Error Distribution'', the intensity of green represents the value of average rendering error of quadtree nodes. Dark green represents large rendering error of the node, while light green represents small rendering error.}
  \label{fig:quadtree-syn}
\end{figure*}

\begin{figure*}[htbp]
  \centering
  \includegraphics[width=\linewidth]{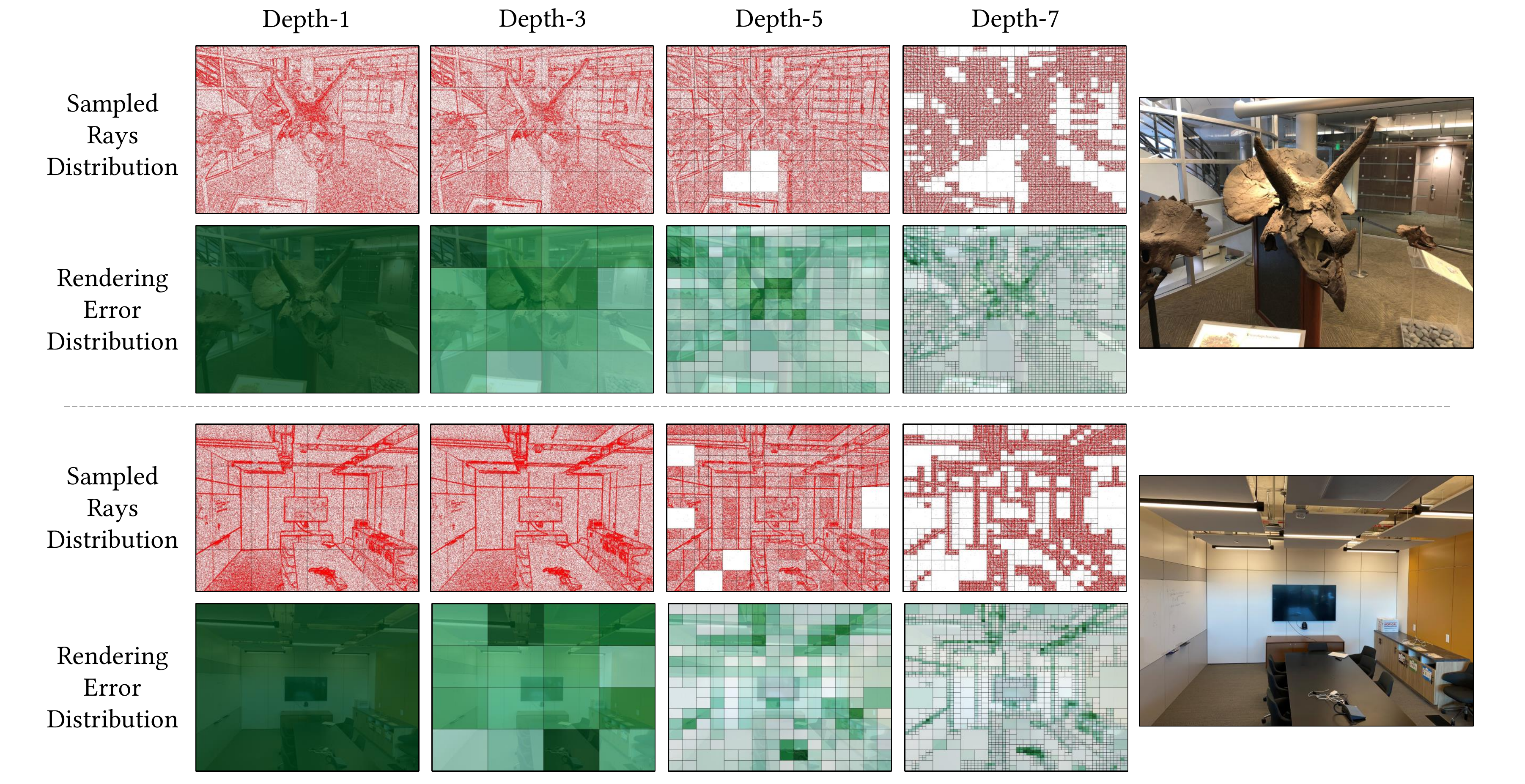}
  \caption{Visualization of sampled rays distribution and rendering error distribution at different quadtree depths. We select ``horns'' and ``room'' in LLFF Dataset~\cite{wu2018light} as examples. The meanings of different colors of the images are the same as Figure~\ref{fig:quadtree-syn}.}
  \label{fig:quadtree-llff}
\end{figure*}

\begin{figure*}[htbp]
  \centering
  \includegraphics[width=\linewidth]{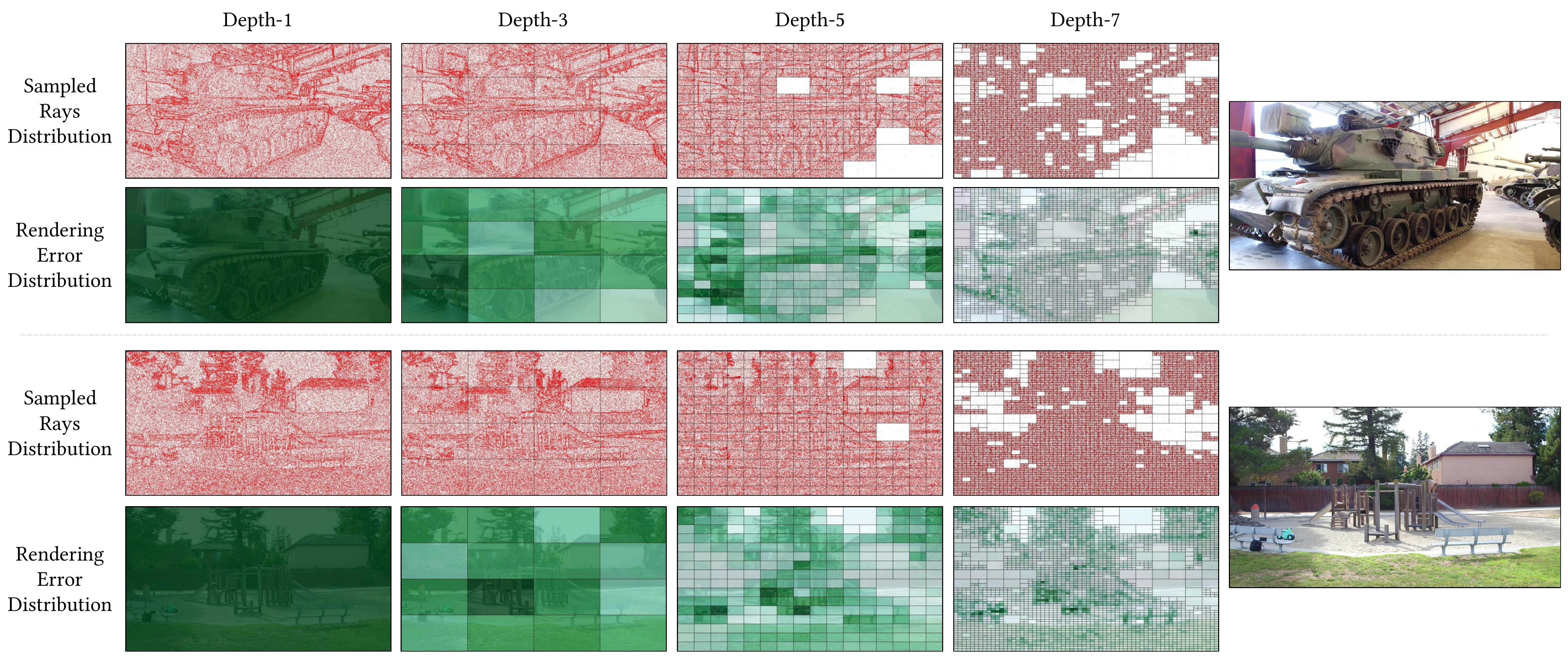}
  \caption{Visualization of sampled rays distribution and rendering error distribution at different quadtree depths. We select ``M60'' and ``Playground'' in Tanks And Temples Dataset~\cite{knapitsch2017tanks} as examples. The meanings of different colors of the images are the same as Figure~\ref{fig:quadtree-syn}.}
  \label{fig:quadtree-tnt}
\end{figure*}

Figures~\ref{fig:quadtree-syn}, \ref{fig:quadtree-llff}, \ref{fig:quadtree-tnt} demonstrate the process of quadtree subdivision. We select several scenes from different datasets and select quadtree depths of 1, 3, 5, 7 for display. In each scene, red images demonstrate the sampled rays distribution and each red point represents a sampled ray. Green images demonstrate the rendering error distribution and the intensity of green represents the value of average rendering error of quadtree nodes. Dark green represents large rendering error on the node while light green represents small rendering error. The distribution of quadtree leaf nodes and rendering error proves the effectiveness of our subdivision strategy. The chosen subdivided leaf nodes are always around the nontrivial areas, so the distribution of leaf nodes will approach the details of the scene or the silhouette of the object. With the increase of subdivision, more leaf nodes are marked and sampled rays are more concentrated in the nontrivial areas, which lead the deep network to learn more details of the scene.

Section~\ref{sec: colorProb} and \ref{sec: quadtree} respectively provide the prior and posterior probability distribution for training rays on images. Prior distribution guides how rays are sampled on each leaf nodes, and posterior distribution determines which redundant blocks are to be discarded. Our method of sampling rays not only improves the rendering effect but also reduces the training time. It should be noted that our method is only available for the training stage. During test stage, the number of the rays to be rendered always equals the number of pixels of the image and can not be reduced, so the rendering procedure can not be accelerated through quadtree subdivision or adaptive sampling.

\subsection{Implementation Details}

\noindent \textbf{All-Pixel Sampling}. To avoid under-fitting, we randomly sample rays from all the pixels in the image instead of using quadtrees for sampling at the last epoch. The color and density in 3D space construct a continuous radiance field. The field will be continuously updated when optimizing the network parameters. Even if the training on some area of an image has converged, the corresponding radiance field of the area may change slowly and slightly in account of the changing of its neighborhood fields. Therefore, the rendered color on marked leaf nodes may gradually deviate from the ground truth color. To avoid this problem, when it comes to the last epoch of training, we just randomly sample rays from the whole image instead of using quadtrees for sampling, where the number of sampled rays is equal to the total number of pixels. Through this way, rays on the marked leaf nodes are trained again at the last epoch to finetune the radiance field. Because the rays pruned by quadtree subdivision have reached the local optimum before, they can converge very quickly at the last epoch of finetuning. We find that the operation of \emph{all-pixel sampling} helps to further improve the rendering accuracy, which will be demonstrated in ablation study below.

\noindent \textbf{Epoch-based Subdivision}. The subdivision of quadtrees starts only when 
loss calculation of all the leaf nodes is complete (i.e. at the end of each epoch).
If the quadtrees are subdivided during an epoch of training, it will be difficult to count the reduction proportion of sampled rays and to control the subdividing process. Following Plenoxels~\cite{fridovich2022plenoxels}, we also generate training rays for all training views before each epoch starts, unlike NeRF~\cite{mildenhall2020nerf} selecting only one view and sampling certain rays from the view at every iteration. At the end of each epoch, we collect the rendering error of rays sampled from each leaf node, which will guide the quadtree subdivision. In this way, we are able to synchronously subdivide all the quadtrees after each epoch ends. 

\noindent \textbf{QuadTree Initialization}. In practice, 
we initially subdivide the quadtrees into 2 or 3 depths at the begin of training. This helps our method to distinguish the trivial and nontrivial areas faster among the quadtree leaf nodes.

%% file: Experiments.tex
\section{Experiments}
\subsection{Experimental Settings}

\noindent \textbf{Dataset.} \ We evaluate our method quantitatively and qualitatively under four widely used benchmarks for novel view synthesis, including Realistic Synthetic 360$^\circ$~\cite{mildenhall2020nerf},  Light Field (LF)~\cite{yucer2016efficient}, Local Light Field Fusion (LLFF)~\cite{mildenhall2019local}, Tanks and Temples (T\&T)~\cite{knapitsch2017tanks}, Real-World 360$^\circ$~\cite{barron2022mip} and one widely used benchmark for multi-view reconstruction, DTU~\cite{jensen2014large}. Synthetic dataset contains pathtraced images of 8 objects that exhibit complex geometry and realistic non-Lambertian materials.
LF dataset is formed by hand-held capturing or a circular boom that rotates around a fixed object. Following NeRF++~\cite{zhang2020nerf++} settings, we use 4 scenes from the LF dataset: Africa, Basket, Ship and Torch. 
LLFF dataset consists of 8 complex real-world scenes captured with roughly forward-facing images. T\&T dataset consists of hand-held 360° captures of large-scale scenes. In our experiments we use two benchmarks of T\&T dataset following the default settings of different baselines. The first one contains four scenes with real-world background, including M60, Playground, Train and Truck. The other one contains five scenes with transparent background, including Barn, Caterpillar, Family, Ignatius and Truck. And Real-World 360° dataset consists of 9 indoor and outdoor scenes, each containing a complex central
object or area and a detailed background. Additionally, DTU dataset contains 15 scenes with a wide variety of materials, appearance and geometry, including challenging cases for reconstruction algorithms, such as non-Lambertian surfaces and thin structures. 

\noindent \textbf{Baselines.} \ To demonstrate our method as a general framework that can work for almost all radiance field based methods, we apply our method to six representative methods, including NeRF~\cite{mildenhall2020nerf}, NeRF++~\cite{zhang2020nerf++}, Plenoxels~\cite{fridovich2022plenoxels}, Instant-NGP~\cite{mueller2022instant}, Mip-NeRF 360~\cite{barron2022mip} and NeuS~\cite{wang2021neus}. Specially, NeRF is the pioneer of neural radiance fields, NeRF++ improves NeRF to represent large-scale unbounded 3D scenes, Plenoxels and Instant-NGP is the latest work on speeding up the learning of radiance fields, Mip-NeRF 360 is the latest work on rendering unbounded scenes and NeuS is a widely adopted method which combines radiance field and neural implicit surfaces to reconstruct multi-view surfaces. Moreover, for fair comparison with Plenoxels, we report the results without the octree pruning in 3D space, since the octree pruning affects the number of rays. We report our improvement in terms of both speed and accuracy. The rendering accuracy is evaluated by PSNR, SSIM and LPIPS, which have been widely adopted by most radiance fields based methods (e.g.~\cite{fridovich2022plenoxels}). Additionally, for DTU dataset, We use Chamfer Distance (CD)~\cite{wang2021neus} to evaluate the quality of reconstructed meshes.

\noindent \textbf{Training details.} \ 
During training, the quadtree is initialized to 2 depths and subdivided every three epochs with a threshold of 1e-3. On each leaf node, 50\% rays are sampled according to the prior distribution and other 50\% rays are randomly sampled. The parameter of random sampling ratio will be discussed in detail in ablation study. Additionally, we randomly sample rays across the whole image at the last epoch to finetune the radiance field. All of the training time of experiments is counted on a single NVIDIA 3090Ti GPU.

\subsection{Results on Synthetic Dataset}
\label{sec: results-on-synthetic-dataset}

\begin{table}[htbp]
  \caption{Quantitative results on Realistic Synthetic 360$^\circ$ dataset. Results are averaged over the 8 synthetic scenes.}
  \label{tab:synthetic}
  \resizebox{\linewidth}{!}{\begin{tabular}{*{6}{c}}
    \toprule
        Methods & Training Time & Reduction & PSNR $\uparrow$ & SSIM $\uparrow$ & LPIPS $\downarrow$  \\
    \midrule
        NeRF~\cite{mildenhall2020nerf} & 22.0h & \multirow{2}{*}{23\%} & 31.14 & 0.950 & 0.067  \\
        Ours(NeRF) & \textbf{16.9h} &  &  \textbf{31.21} & \textbf{0.951} & \textbf{0.066}  \\
    \cmidrule{1-6}
        Plenoxels~\cite{fridovich2022plenoxels} & 25.6min & \multirow{2}{*}{23\%} & \textbf{30.87} & 0.959 & \textbf{0.051}  \\
        Ours(Plenoxels) & \textbf{19.6min} &  & 30.84 & \textbf{0.960} & 0.054  \\
    \cmidrule{1-6}
        Instant~\cite{mueller2022instant} & 264s & \multirow{2}{*}{18\%} & 32.04 & 0.961 & 0.045  \\
        Ours(Instant) & \textbf{217s} &  & \textbf{32.44} & \textbf{0.962} & \textbf{0.043}  \\
  \bottomrule
\end{tabular}}
\end{table}

\begin{figure}[htbp]
  \centering
  \includegraphics[width=\linewidth]{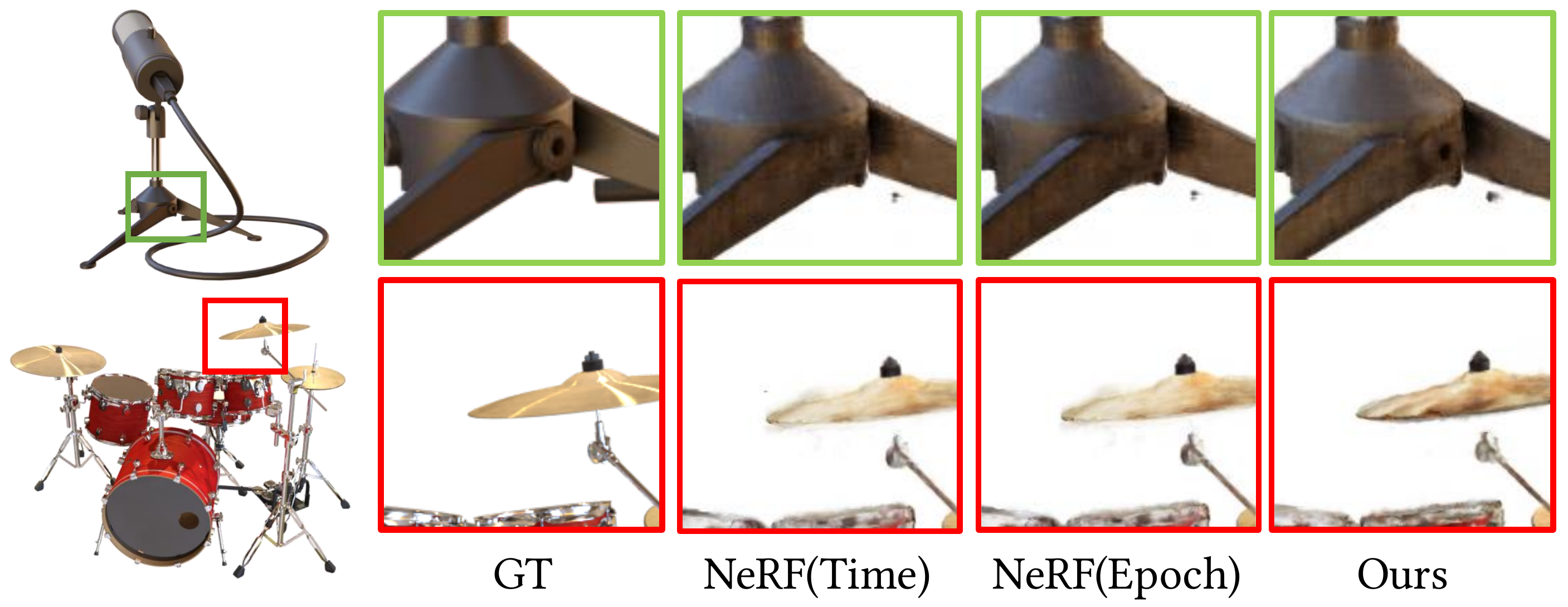}
  \caption{Qualitative comparison of our method vs. NeRF on the Realisitic Synthetic 360$^\circ$ dataset~\cite{mildenhall2020nerf}.}
  \label{fig:ours-vs-nerf}
\end{figure}

Table~\ref{tab:synthetic} shows the quantitative comparisons with NeRF~\cite{mildenhall2020nerf}, Plenoxels~\cite{fridovich2022plenoxels} and Instant-NGP~\cite{mueller2022instant} under Realistic Synthetic 360$^\circ$ dataset. We report our results upon NeRF, Plenoxels and Instant-NGP, as denoted by ``Ours(NeRF)'', ``Ours(Plenoxels)'' and ``Ours(Instant)'', respectively. In Table~\ref{tab:synthetic}, the bold items indicate better results in the numerical comparison, which is the same setting as used below. Our method can save 23\% training time (from 22 hours to 16.9 hours) for NeRF and 23\% training time (from 25.6 minutes to 19.6 minutes) for Plenoxels on synthetic dataset, where we achieve comparable results to Plenoxels. Althogh Instant-NGP is the state-of-the-art work on accelerating NeRF training and is able to train a radiance field in less than 5 minues, our method upon Instant-NGP further saves 18\% training time (from 264s to 217s) and achieves better performance than Instant-NGP.
Figure~\ref{fig:ours-vs-nerf} provides some qualitative results between NeRF and our method, where more details are demonstrated in our results. We compare with NeRF in two training settings in the visualization results. In the first setting, we train NeRF in the same time as ours, and in the second setting, we train NeRF in the same epochs as ours, as denoted by ``NeRF(Time)'' and ``NeRF(Epoch)'', respectively. The two settings of other methods below are the same as those of NeRF.

We noticed that our numerical results are merely a little bit better than baselines in Table~\ref{tab:synthetic}, \ref{tab:LLFF dataset}, \ref{tab:TNT dataset}. However, the marginal improvements are caused by the optimization of NeRF. A remarkable phenomenon of NeRF-based methods is that the accuracy increases rapidly at the beginning of training, while increasing extremely slowly after some epochs. For example, PSNR of NeRF in synthetic dataset increases less than 1.0 at the last 5 hours, which is demonstrated in (b) and (c) of Figure~\ref{fig:teaser}. Additionally, we provide some qualitative results between our method and baseline methods, in which our method shows obvious advantages on the scene details, as shown in Figure~\ref{fig:ours-vs-nerf} - \ref{fig:ours-vs-nerf++-Tanks} in the paper.

\begin{table*}[htbp]
  \centering
  \caption{The comparison of PSNR and training time between Plenoxels~\cite{fridovich2022plenoxels} and our method after each epoch.}
  \label{tab:10epoch compare}
  \begin{tabular}{*{12}{c}}
    \toprule
         & Epoch & 0 & 1 & 2 & 3 & 4 & 5 & 6 & 7 & 8 & 9  \\
    \midrule
        \multirow{3}{*}{PSNR $\uparrow$} & Plenoxels~\cite{fridovich2022plenoxels} & 10.427 & 30.198 & 31.218 & 31.696 & 31.965 & 32.140 & 32.245 & 32.327 & 32.373 & 32.405  \\
         & Ours & 10.427 & 30.336 & 31.317 & 31.755 & 32.010 & 32.201 & 32.315 & 32.375 & 32.449 & 32.464  \\
         & Improvements & 0.000 & \textbf{+0.138} & \textbf{+0.099} & \textbf{+0.059} & \textbf{+0.045} & \textbf{+0.061} & \textbf{+0.070} & \textbf{+0.048} & \textbf{+0.076} & \textbf{+0.059}  \\
    \cmidrule{1-12}
        \multirow{3}{*}{Time(s) $\downarrow$} & Plenoxels~\cite{fridovich2022plenoxels} & 0 & 176.08 & 325.85 & 473.6 & 619.34 & 764.34 & 909.64 & 1053.4 & 1197.67 & 1342.52  \\
         & Ours & 0 & 185.65 & 346.21 & 465.69 & 590.4 & 682.89 & 789.62 & 875.16 & 976.27 & 1049.91  \\
         & Improvements & 0.00 & +9.57 & +20.36 & \textbf{-7.91} & \textbf{-28.94} & \textbf{-81.45} & \textbf{-120.02} & \textbf{-178.24} & \textbf{-221.40} & \textbf{-292.61} \\
  \bottomrule
\end{tabular}
\end{table*}

We also conduct an experiment to illustrate the above phenomenon. We train “Plenoxels” and “Ours(Plenoxels)” on the “Lego” scene from synthetic dataset for 10 epochs. We report the PSNR and time after each one of the first 10 epochs. The initial depth of quadtrees is 3 and the quadtrees are subdivided every 2 epochs (same as the settings in paper). The PSNR (the larger the better) and the training time of some epochs are listed in Table~\ref{tab:10epoch compare}. As shown in the table, at the beginning of training, the PSNR of both our method and Plenoxels increases rapidly after the first epoch. Benefiting from our context-based probability sampling strategy, we achieved larger PSNR improvements over Plenoxels after each epoch. Additionally, as the training progresses, our PSNR improvements over Plenoxels become smaller. It is because that the bottleneck capacity of Plenoxels restricts the upper limit of PSNR. At the same time, our training speed shows significant improvements because of the quadtree subdivision strategy (the tiny increase of training time at the beginning is because of sampling and subdividing). Therefore, these results demonstrate that our method can achieve a balance of efficiency and effectiveness, i.e. better performance improvements at the early training stages or larger training speed improvements at the later training stages.

\subsection{Results on Real-World Dataset}
\label{sec: results-on-real-world-dataset}

\begin{table}[htbp]
  \caption{Quantitative results on Forward-Facing dataset. Results are averaged over the 8 scenes.}
  \label{tab:LLFF dataset}
  \resizebox{\linewidth}{!}{\begin{tabular}{*{6}{c}}
    \toprule
        Methods & Training Time & Reduction & PSNR $\uparrow$ & SSIM $\uparrow$ & LPIPS $\downarrow$  \\
    \midrule
        NeRF~\cite{mildenhall2020nerf} & 4.4h & \multirow{2}{*}{16\%} & \textbf{25.85} & 0.782 & \textbf{0.288}  \\
        Ours(NeRF) & \textbf{3.8h} &  & 25.79 & \textbf{0.784} & 0.294  \\
    \cmidrule{1-6}
        Plenoxels~\cite{fridovich2022plenoxels} & 66.6min & \multirow{2}{*}{39\%} & 24.60 & \textbf{0.768} & \textbf{0.316}  \\
        Ours(Plenoxels) & \textbf{40.4min} &  & \textbf{25.12} & 0.767 & 0.322  \\
    \cmidrule{1-6}
        Instant~\cite{mueller2022instant} & 350s & \multirow{2}{*}{15\%} & 26.03 & 0.810 & 0.265  \\
        Ours(Instant) & \textbf{296s} &  & \textbf{26.39} & \textbf{0.812} & \textbf{0.264}  \\
  \bottomrule
\end{tabular}}
\end{table}

\begin{figure}[htbp]
  \centering
  \includegraphics[width=\linewidth]{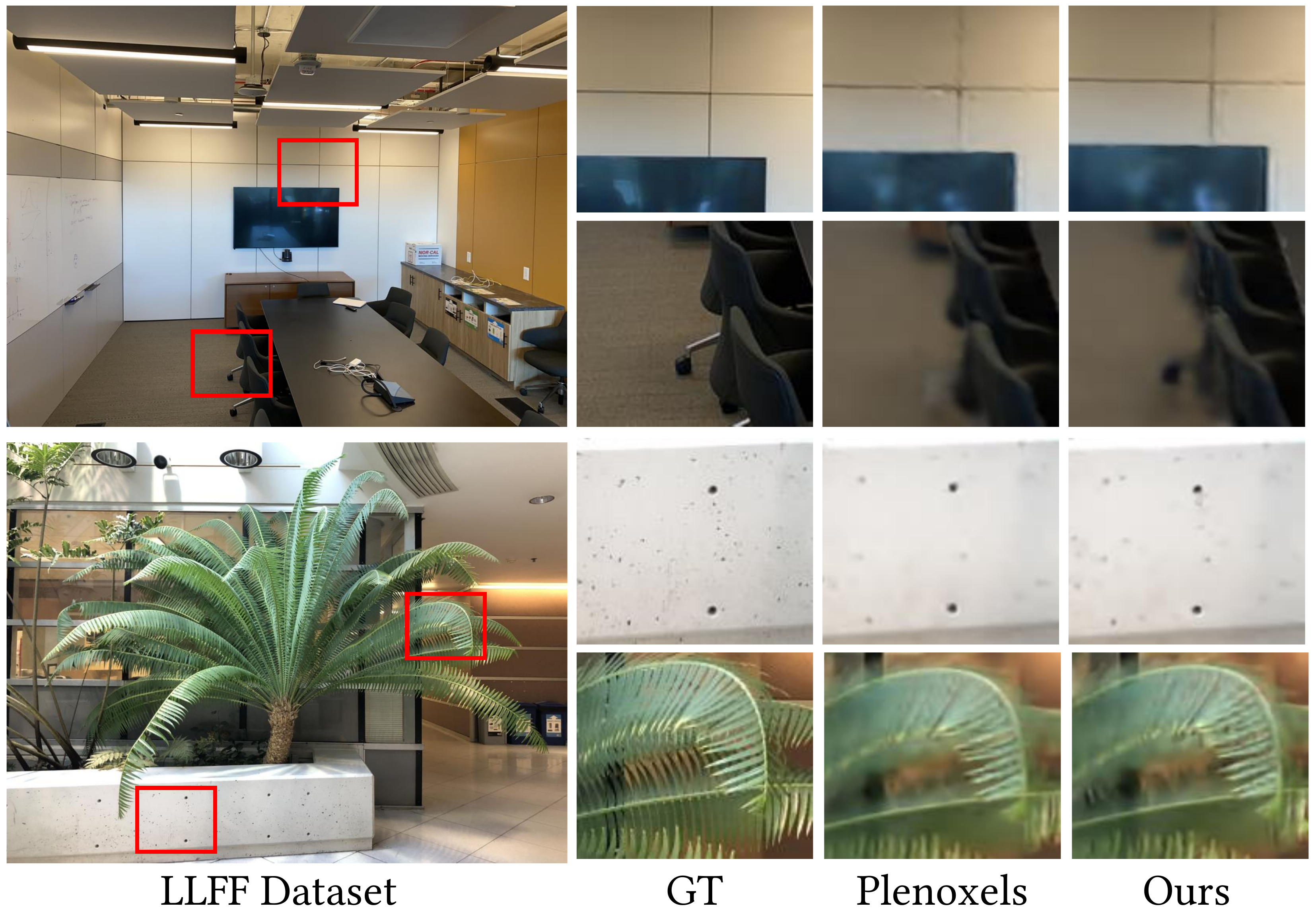}
  \caption{Qualitative comparison of our method vs. Plenoxels on the LLFF dataset~\cite{mildenhall2019local}. Take cracks in walls, spots in flower beds for examples.}
  \label{fig:ours-vs-plenoxels}
\end{figure}

\begin{table}[htbp]
  \caption{Quantitative results on Light Field dataset. Results are averaged over the 5 scenes.}
  \label{tab:LF dataset}
  \begin{tabular}{*{5}{c}}
    \toprule
        Methods & Training Time & PSNR $\uparrow$ & SSIM $\uparrow$ & LPIPS $\downarrow$  \\
    \midrule
        NeRF++(Time) & 6.7h & 24.59 & 0.798 & 0.331  \\
        NeRF++(Epoch) & 7.6h & 25.14 & 0.811 & 0.318  \\
    \cmidrule{1-5}
        Ours(NeRF++) & 6.7h & 25.06 & 0.807 & 0.324  \\
  \bottomrule
\end{tabular}
\end{table}

\begin{figure*}[htbp]
  \centering
  \includegraphics[width=\linewidth]{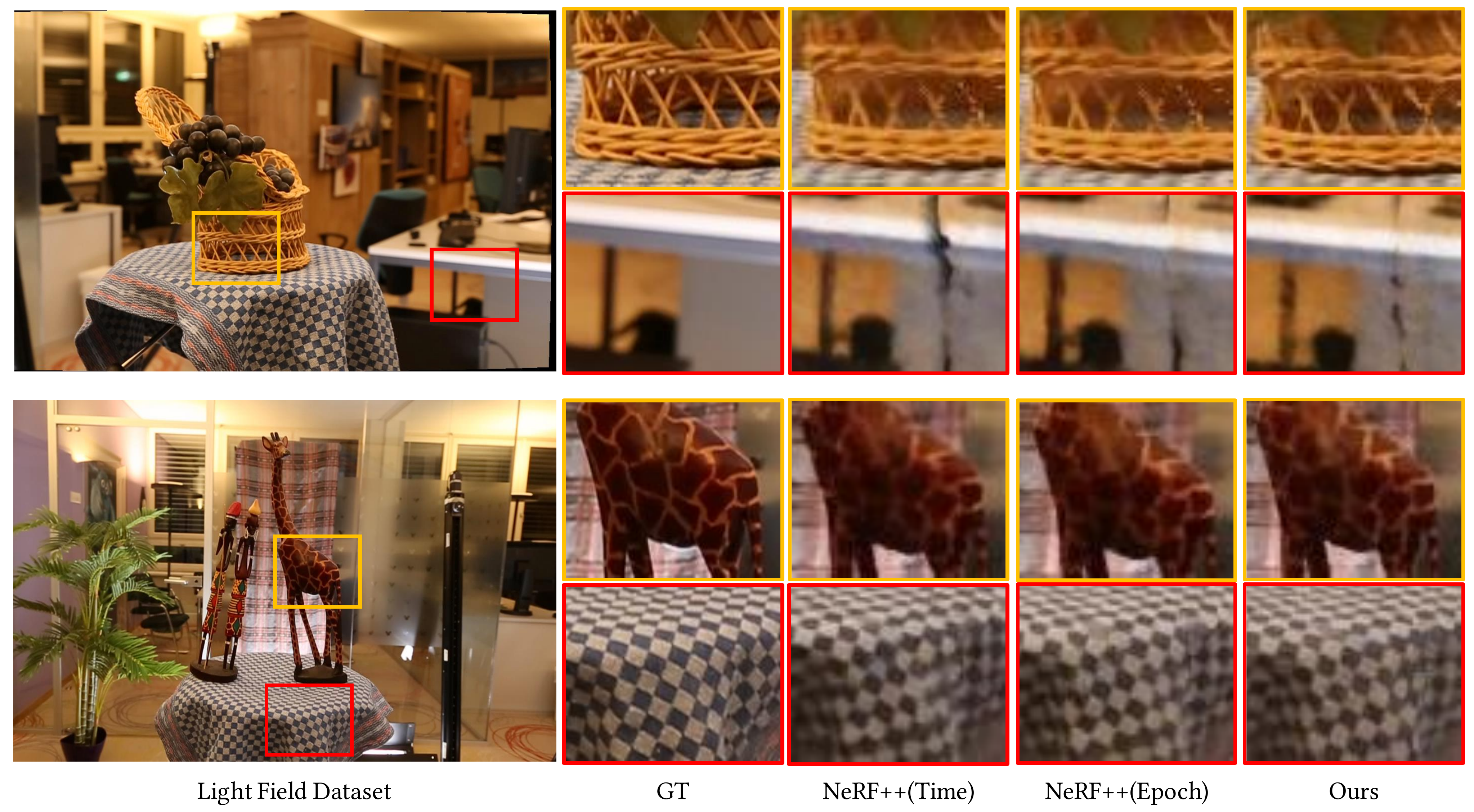}
  \caption{Qualitative comparison of our method vs. NeRF++~\cite{zhang2020nerf++} on the Light Field dataset~\cite{yucer2016efficient}. Comparing with NeRF++, our method is faster and captures the scene details well, taking reticular formation of basket, patterns on giraffes for examples.}
  \label{fig:ours-vs-nerf++-LF}
\end{figure*}

\begin{table}[htbp]
\caption{Quantitative results on Tanks and Temples dataset\cite{knapitsch2017tanks}. Results are averaged over all the scenes.}
  \label{tab:TNT dataset}
  \resizebox{\linewidth}{!}{\begin{tabular}{*{6}{c}}
    \toprule
        Methods & Training Time & Reduction & PSNR $\uparrow$ & SSIM $\uparrow$ & LPIPS $\downarrow$  \\
    \midrule
        NeRF++~\cite{zhang2020nerf++} & 7.7h & \multirow{2}{*}{18\%} & \textbf{20.42} & 0.659 & \textbf{0.417}  \\
        Ours(NeRF++) & \textbf{6.3h} &  & 20.39 & \textbf{0.663} & 0.421  \\
    \cmidrule{1-6}
        Plenoxels~\cite{fridovich2022plenoxels} & 41.0min & \multirow{2}{*}{20\%} & 19.88 & 0.669 & 0.473  \\
        Ours(Plenoxels) & \textbf{32.7min} &  & \textbf{19.98} & \textbf{0.671} & \textbf{0.469}  \\
    \cmidrule{1-6}
        Instant~\cite{mueller2022instant} & 354s & \multirow{2}{*}{19\%} & 27.91 & 0.832 & 0.141   \\
        Ours(Instant) & \textbf{285s} &  & \textbf{28.25} & \textbf{0.837} & \textbf{0.132}  \\
  \bottomrule
\end{tabular}}
\end{table}

\begin{table}[htbp]
\caption{Quantitative results on Real-World 360$^\circ$ Dataset\cite{barron2022mip}. Results are averaged over the 7 scenes.}
  \label{tab:real-world 360 dataset}
  \resizebox{\linewidth}{!}{\begin{tabular}{*{6}{c}}
    \toprule
        Methods & Training Time & Reduction & PSNR $\uparrow$ & SSIM $\uparrow$ & LPIPS $\downarrow$  \\
    \midrule
        Mip-NeRF 360~\cite{barron2022mip} & 41.5h & \multirow{2}{*}{23\%} & 26.45 & 0.785 & 0.242  \\
        Ours(Mip-NeRF 360) & \textbf{31.9h} &  & \textbf{26.89} & \textbf{0.787} & \textbf{0.240}  \\
  \bottomrule
\end{tabular}}
\end{table}

\begin{table}[htbp]
  \caption{Quantitative results on DTU Dataset\cite{jensen2014large}. Results are averaged over the 15 scenes.}
  \label{tab:dtu dataset}
  \resizebox{\linewidth}{!}{\begin{tabular}{*{5}{c}}
    \toprule
        Methods & Training Time & Reduction & PSNR $\uparrow$ & CD $\downarrow$   \\
    \midrule
        NeuS~\cite{wang2021neus} & 9.2h & \multirow{2}{*}{21\%} & 31.97 & 0.87   \\
        Ours(NeuS) & \textbf{7.3h} &  & \textbf{33.41} & \textbf{0.73}   \\
   \bottomrule
  \end{tabular}}
\end{table}

\begin{figure*}[htbp]
  \centering
  \includegraphics[width=\linewidth]{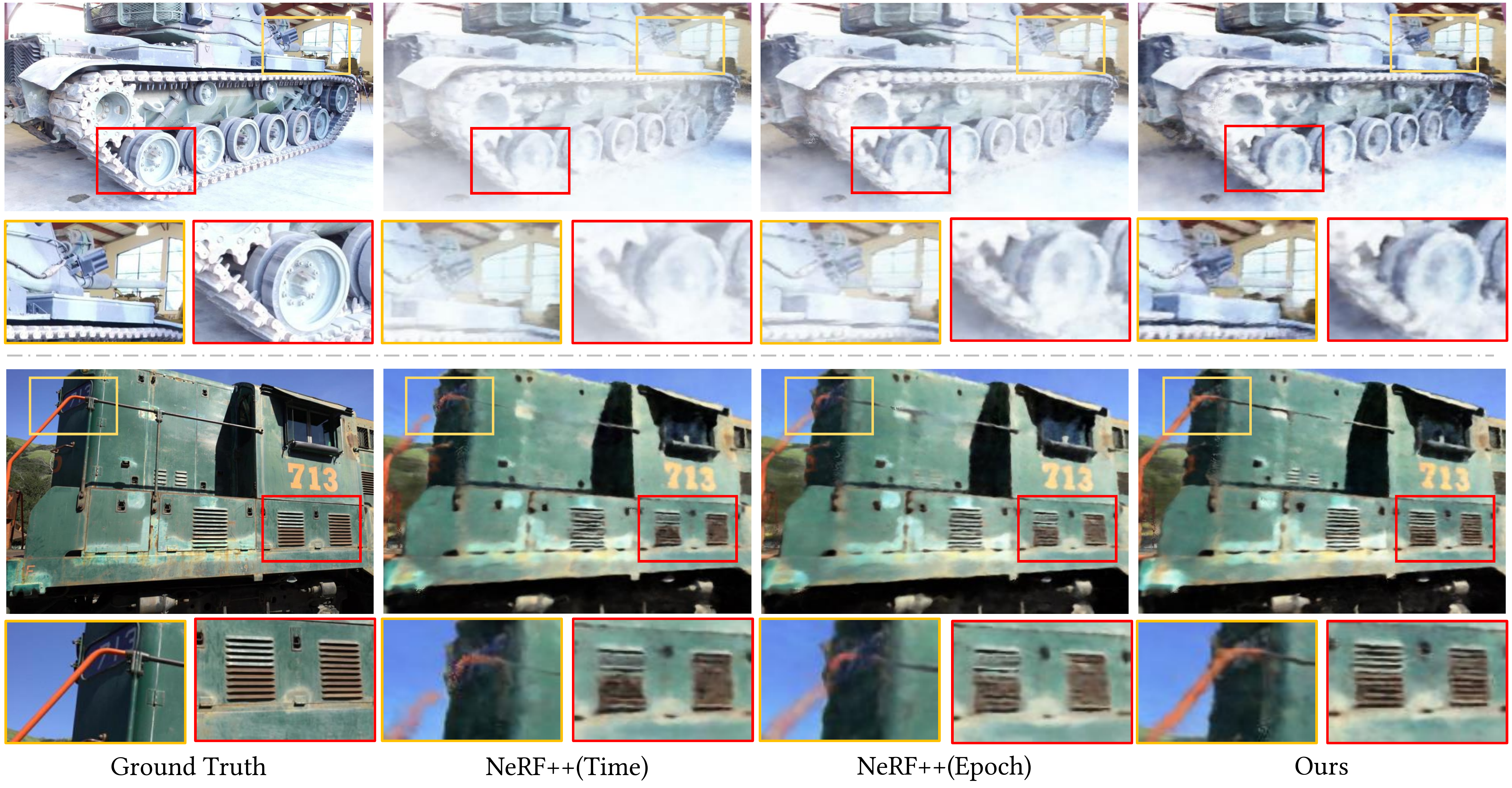}
  \caption{Qualitative comparison of our method vs. NeRF++~\cite{zhang2020nerf++} on the Tanks and Temples dataset~\cite{knapitsch2017tanks}. Comparing with NeRF++, our method achieves both faster speed and better visual performance, taking gear of tank, exhaust fan of train for examples.}
  \label{fig:ours-vs-nerf++-Tanks}
\end{figure*}

\begin{figure*}[htbp]
  \centering
  \includegraphics[width=\linewidth]{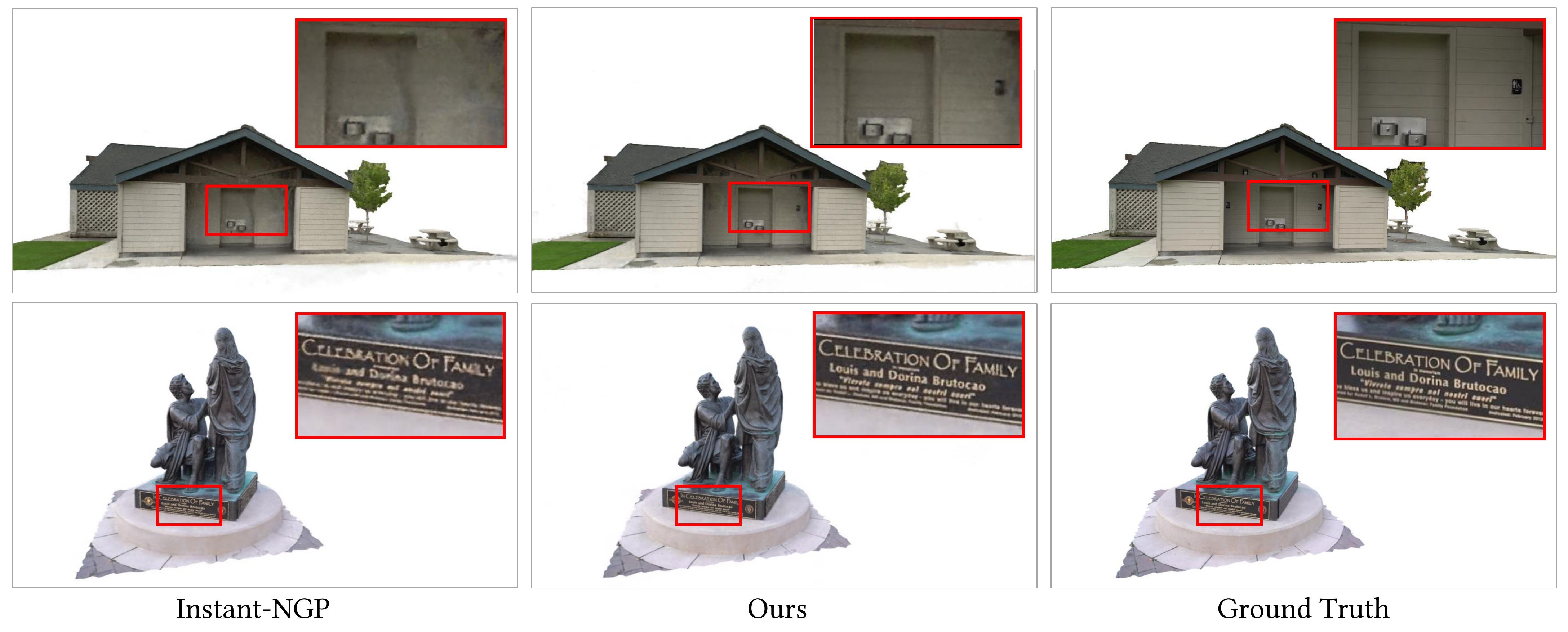}
  \caption{Qualitative comparison of our method vs. Instant-NGP~\cite{mueller2022instant} on the Tanks and Temples dataset~\cite{knapitsch2017tanks}. Our method further improves the training speed and the performance of Instat-NGP, which is the state-of-the-art among NeRF accelerating methods.}
  \label{fig:ours-vs-instant}
\end{figure*}

\begin{figure*}[htbp]
  \centering
  \includegraphics[width=\linewidth]{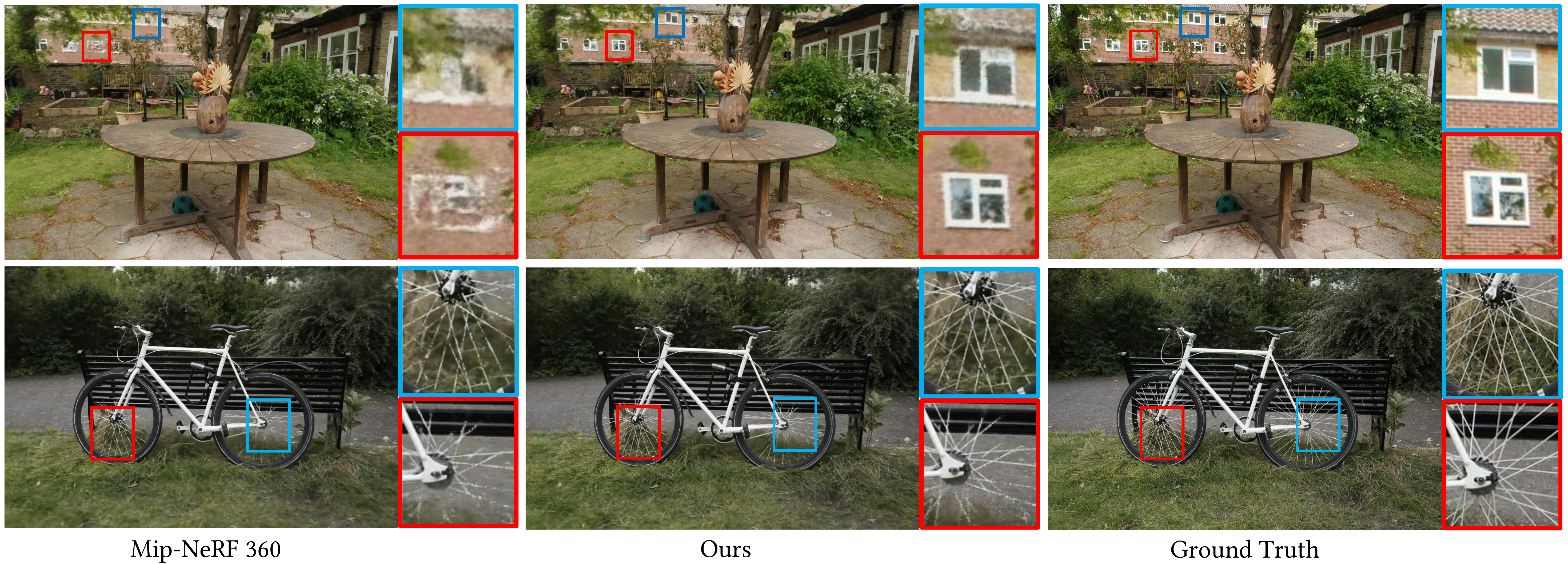}
  \caption{Qualitative comparison of our method vs. Mip-NeRF 360~\cite{barron2022mip} on the Real-World 360$^\circ$~\cite{barron2022mip}. Our method improves both the training speed and the performance of Mip-NeRF 360, which is the state-of-the-art among Realistic 360$^\circ$ rendering methods.}
  \label{fig:ours-vs-mipnerf}
\end{figure*}

\begin{figure*}[htbp]
  \centering
  \includegraphics[width=\linewidth]{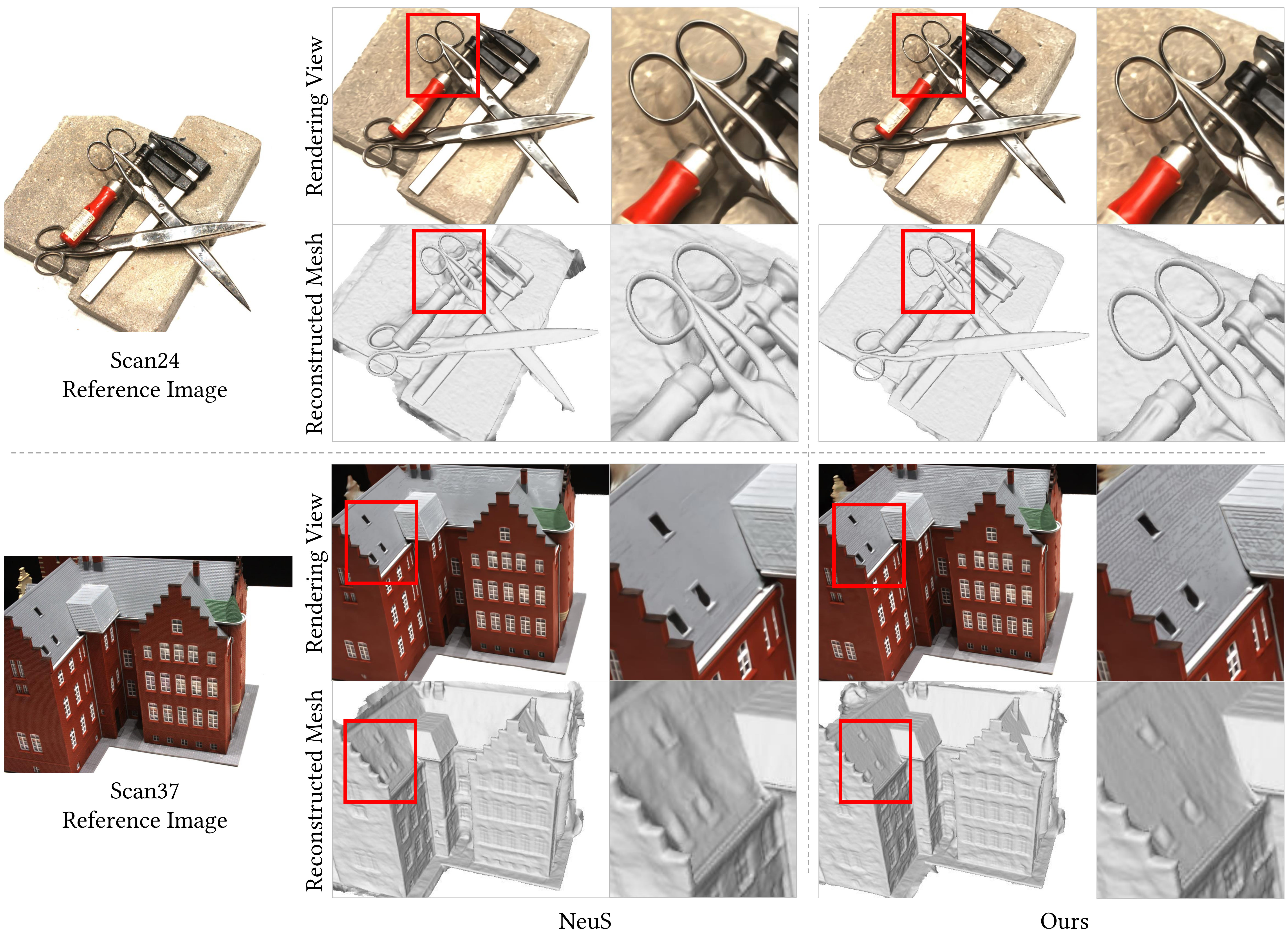}
  \caption{Qualitative comparison of our method vs. NeuS~\cite{wang2021neus} on DTU dataset~\cite{jensen2014large}. For each scene, we display the reference view, rendering view and the reconstructed mesh between our method and NeuS. The results show that our method can improve both the rendering quality and the reconstruction details of NeuS.}
  \label{fig:ours-vs-neus}
\end{figure*}

We then evaluate our method under Forward-Facing (LLFF) dataset by comparing with NeRF, Plenoxels and Instant-NGP. As shown in Table~\ref{tab:LLFF dataset}, with the help of our framework, we can speed up NeRF by 16\%  (from 4.4 hours to 3.8 hours), Plenoxels by 39\% (from 66.6 minutes to 40.4 minutes) and Instant-NGP by 15\% (from 350s to 296s), respectively. Figure~\ref{fig:ours-vs-plenoxels} provides some qualitative results between Plenoxels and our method. The results show that our method can predict more details such as the office chair leg, the cracks of the walls, the spots on the flower beds and the gap between leaves in the novel synthetic views.

In Table \ref{tab:LF dataset}, we further report numerical comparison with NeRF++ under Light Field (LF) dataset. Note that almost all follow-up approaches \cite{wu2018light, Turki_2022_CVPR, shen2022conditional} reporting results under LF dataset use different dataset splits, thus their results are not comparable with ours. Hence, we only list the results of NeRF++ and ours on this dataset. We report the result of NeRF++ by training it for the same time or the same number of epochs as ours, as denoted by ``NeRF++(Time)'' and ``NeRF++(Epoch)'', respectively. The settings of ``Time'' and ``Epoch'' are the same as the ones in Figure~\ref{fig:ours-vs-nerf}. With the same training time, our method achieves better accuracy than NeRF++, while NeRF++ achieves comparable results to ours with more time cost. We further highlight our advantage in visual comparison in Figure~\ref{fig:ours-vs-nerf++-LF}. Our method can reveal more fine texture and structures in novel views.

We report numerical comparisons under Tanks and Temples dataset in Table~\ref{tab:TNT dataset}, which further demonstrates that our method is a general framework for most radiance fields based methods. Our methods upon NeRF++, Plenoxels and Instant-NGP can significantly speed up their training without scarifying accuracy. 
Figure~\ref{fig:ours-vs-instant} and Figure~\ref{fig:ours-vs-nerf++-Tanks} are visual comparison with NeRF++ and Instant-NGP and our method under T\&T dataset. Note that we use the benchmark of 4 scenes with real background for NeRF++ and Plenoxels while use the benchmark of 5 scenes with transparent background for Instant-NGP, following the same settings of the baselines. The results show that our method can show more details with fewer artifacts.

We then evaluate our method by comparing with Mip-NeRF 360~\cite{barron2022mip} under Real-World 360° dataset~\cite{barron2022mip}, as reported in Table~\ref{tab:real-world 360 dataset}. Mip-NeRF 360 is the state-of-the-art work on rendering unbounded scenes and our method further speeds up Mip-NeRF 360 by 23\% (from 41.5h to 31.9h) and achieves better numerical performance. Figure~\ref{fig:ours-vs-mipnerf} provides some visual examples between our method and Mip-NeRF 360, where our method shows significant advantages in scene details, such as the windows in the distance.

\subsection{Results on Multi-view-stereo Dataset}

In Sections~\ref{sec: results-on-synthetic-dataset} and \ref{sec: results-on-real-world-dataset}, we evaluate our method by comparing with several representative methods which focus on novel view synthesis. Our method is able to reduce the training time from 15\% to 40\% and achieve comparable numerical results under different baselines. However, our method is not limited for novel view synthesis task. To further demonstrate that our method is a general strategy for almost all radiance fields based methods, we integrate our method with NeuS~\cite{wang2021neus}, which combines radiance fields and neural implicit surfaces and is designed for multi-view reconstruction task. We evaluate our method under all the 15 scenes on DTU dataset~\cite{jensen2014large} and report the quality of rendering views (evaluated by PSNR) and reconstructed meshes (evaluated by CD) in Table~\ref{tab:dtu dataset}. The results show that our methods reduce 21\% training time (from 9.2h to 7.3h) for NeuS and achieves better performance on both rendering views and reconstructed meshes. Figure~\ref{fig:ours-vs-neus} provides visual comparison between our method and NeuS. With the help of probability sampling and quadtree subdivision, our method is able to render high-fidelity images with smooth clean surfaces, taking scissors handles in scan24 and roofs in scan37 as examples.

\subsection{Ablation Study \label{sec:ablation}}

\begin{figure}[htbp]
  \centering
  \includegraphics[width=\linewidth]{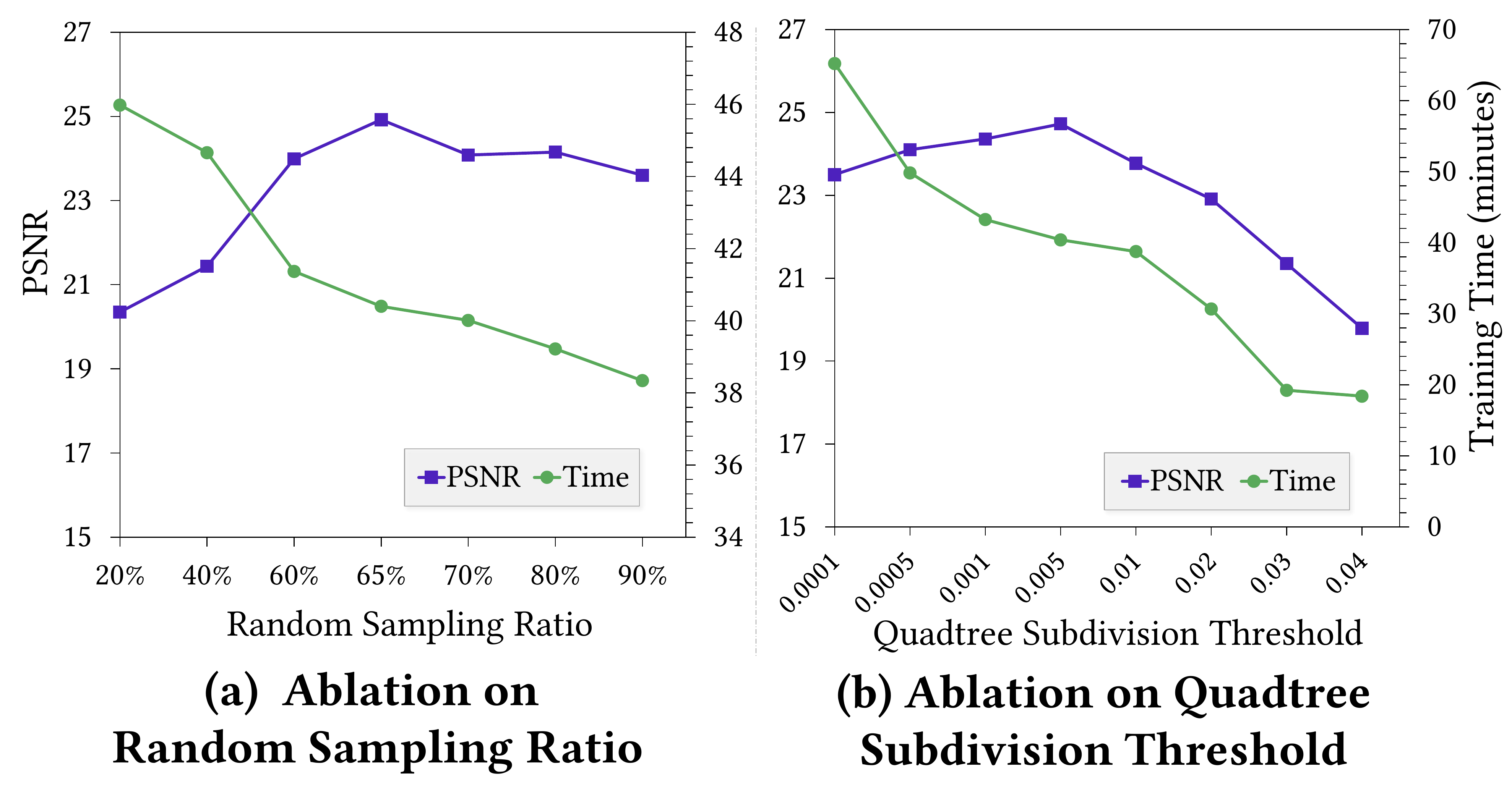}
  \caption{Ablation study on random sampling ratio and quadtree subdivision thresholds. }
  \label{fig:ablation-ratio}
\end{figure}

\begin{figure}[htbp]
  \centering
  \includegraphics[width=\linewidth]{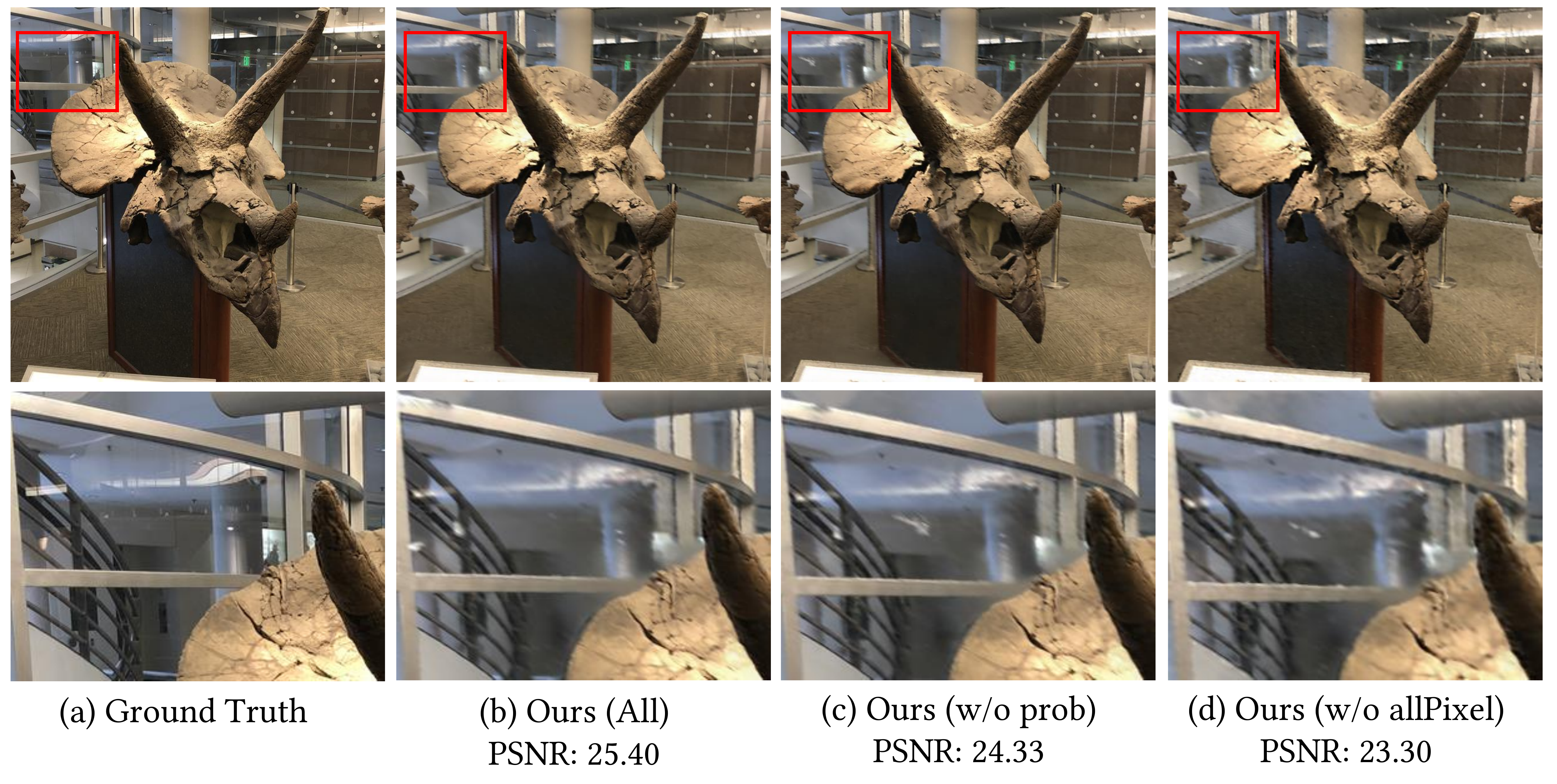}
  \caption{Qualitative comparison of ablation study on each module of our method.}
  \label{fig:ablation-visual-compare}
\end{figure}

\begin{figure*}[htbp]
  \centering
  \includegraphics[width=\linewidth]{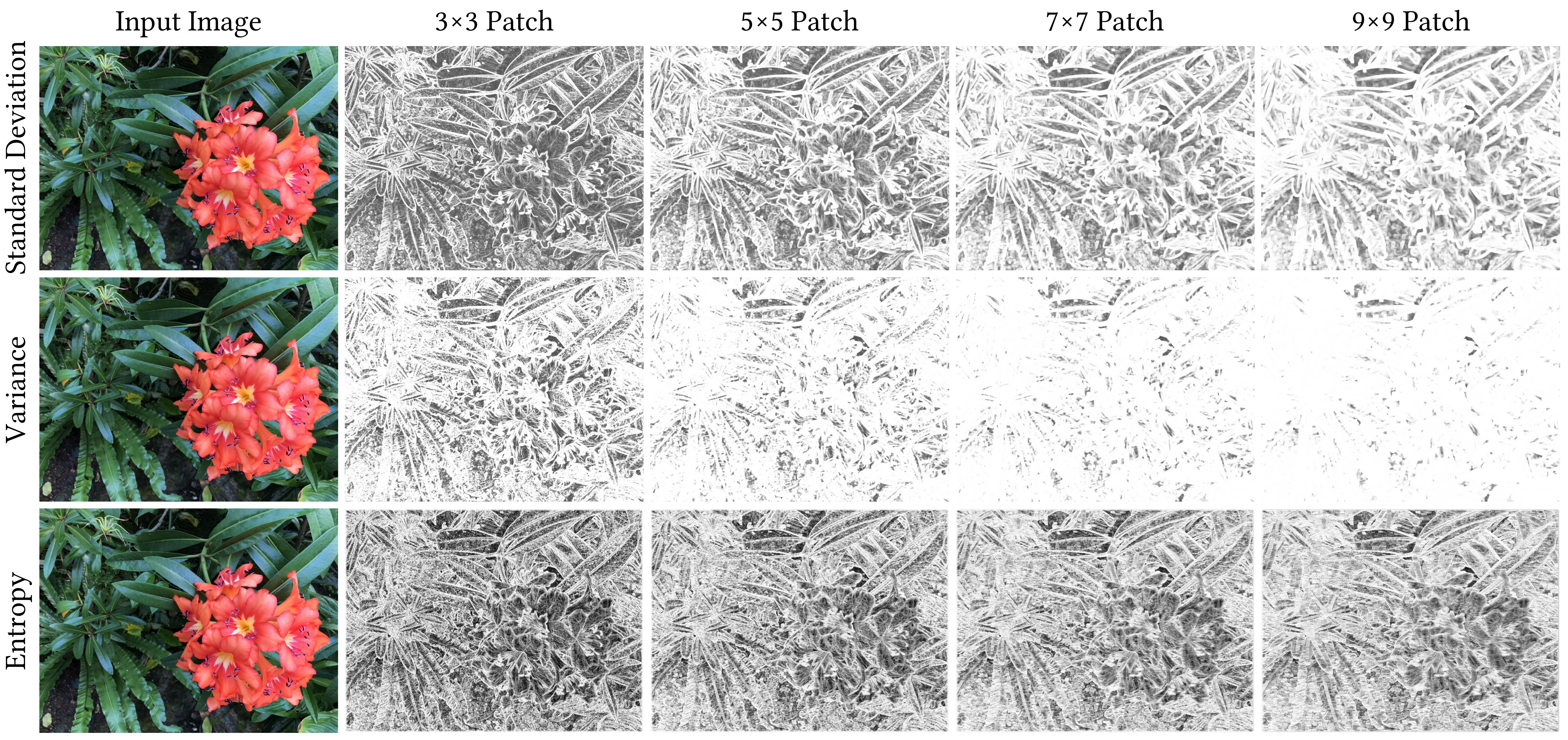}
  \caption{Visualization of different choices of measuring context based probability distribution. Each line represents patches of different scales, and each column represents different approaches of calculating context information.}
  \label{fig:ablation-patch-visual}
\end{figure*}

\begin{figure}[htbp]
  \centering
  \includegraphics[width=\linewidth]{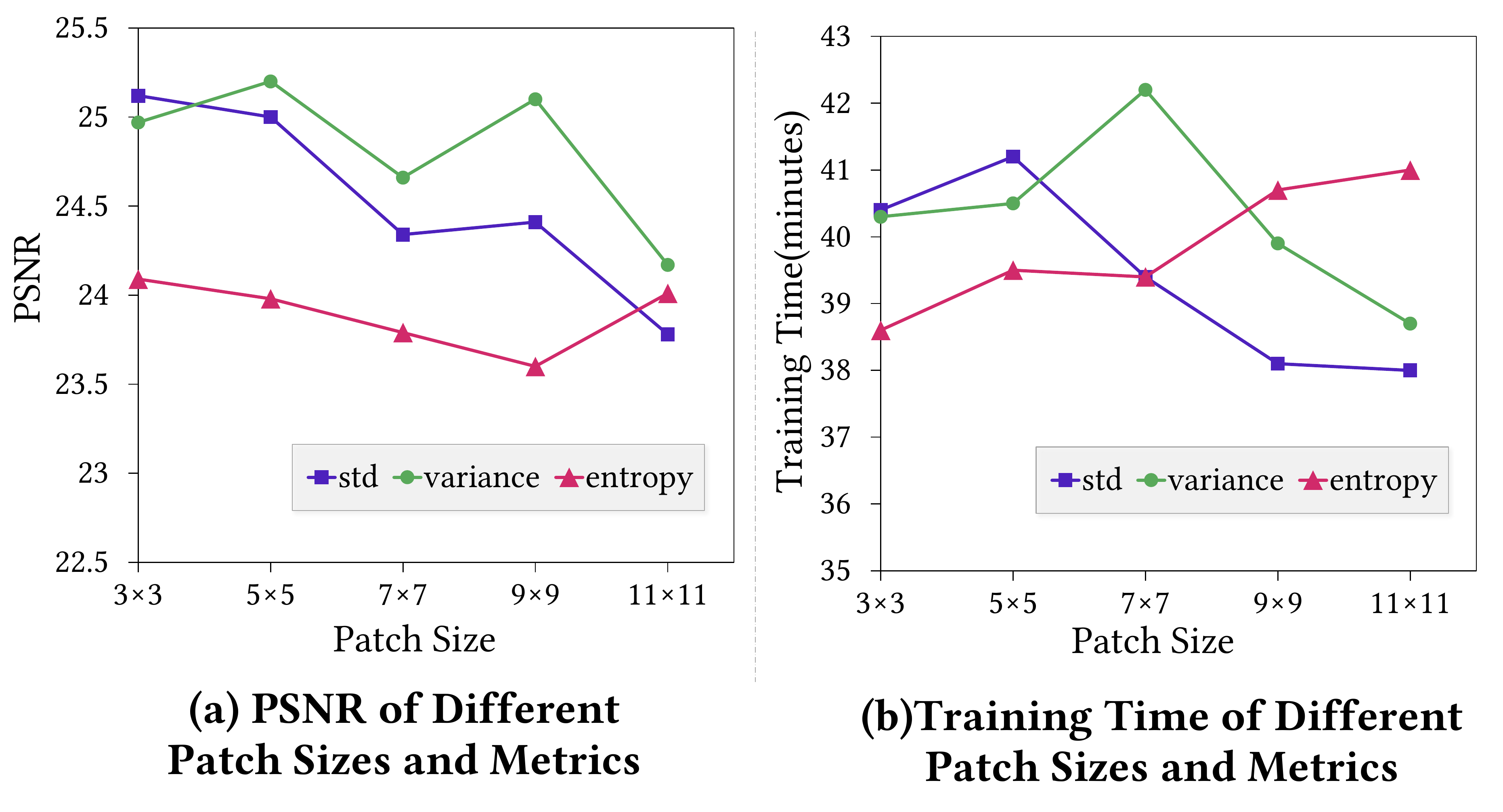}
  \caption{Ablation study on different sizes of patches and approaches of context information metrics.}
  \label{fig:ablation-patch-curse}
\end{figure}

\begin{table}[htbp]
  \centering
  \caption{Ablation study on various components of our method.}
  \label{tab:ablation-quadtree}
  \begin{tabular}{*{3}{c}}
    \toprule
        Methods & Training Time & PSNR $\uparrow$  \\
    \midrule
        Plenoxels         & 66.6min & 24.60   \\
        Ours              & 40.4min & 25.12   \\
        Ours w/o quadtree & 71.5min & \textbf{25.31}   \\
        Ours w/o prob     & 38.4min & 23.97   \\
        Ours w/o allPixel & \textbf{37.8min} & 23.01   \\
        Ours w/o random   & 42.8min & 23.39   \\
  \bottomrule
\end{tabular}
\end{table}

\noindent \textbf{Ratio of random sampling.} We find that adding a certain ratio of random sampling in probability distribution sampling is helpful to learn the radiance field better. It is because that the context based probability distribution may ignore some trivial but important areas on the image, such as the areas of smooth surfaces. Moreover, random sampling helps to pay more attention to the trivial areas, which usually have small color change, so this strategy intuitively plays the same role as threshold $s$ in Eq.~(\ref{eq: probability-normalization}). In this part of ablation study, we freeze quadtree settings and report the correlation between PSNR, training time and random sampling ratio on Forward-Facing dataset~\cite{mildenhall2019local} based on Plenoxels~\cite{fridovich2022plenoxels}. The metrics are averaged across the eight scenes of the dataset, as shown in Figure~\ref{fig:ablation-ratio} (a). With the increase of ratio from 20\% to 90\%, a peak value occurs to PSNR, while the training time keeps decreasing. This is because that with the increase of random sampling ratio, our method samples more rays in trivial areas, which (1) changes the ability of learning details, leading to a wave of PSNR, and (2) results in leaf nodes being marked more easily, leading to a decline of training time. The results are not sensitive with large random sampling ratio ($>$60\%) while suffer from low random sampling ratio. It is because that low random sampling ratio leads to a lack of learning in trivial areas, while the metrics do not reflect subtle differences on the ratios with high random sampling ratio.

\noindent \textbf{Threshold of Quadtree Subdivision.} The threshold $a$ introduced in Section~\ref{sec: quadtree} is an important hyper-parameter in our experiments. The leaf node with average rendering error larger than $a$ will be subdivided into four smaller leaf nodes, while the leaf node with error smaller than $a$ will be marked and will not be subdivided anymore. Therefore, the threshold $a$ greatly affects the training speed and effect. In this part of ablation study, we freeze the random sampling ratio as 65\% and report the correlation between subdivision threshold $a$, PSNR and training time on the same dataset as the previous ablation experiment based on Plenoxels~\cite{fridovich2022plenoxels}, as shown in Figure~\ref{fig:ablation-ratio} (b). With the increasing of the threshold from 0.0001 to 0.04, the training time keeps decreasing, because larger threshold results in fewer leaf nodes to be subdivided and fewer rays to be trained. However, the metrics are not monotonous with the threshold. If the threshold is too small, most leaf nodes are subdivided, so the rays are sampled on very small leaf nodes, which is similar to random sampling, losing the advantage of our context-based sampling strategy. On the other hand, if the threshold is too large, a leaf node may be marked to be not subdivided anymore although the training on this node has not converged yet.

\noindent \textbf{Metrics of context based probability distribution.} In Eq.~\ref{eq:std}, we propose using the standard deviation of the color in $3\times 3$ patch to represent the probability density of a given pixel. In this part, we compare the performance of different sizes of patches and different kinds of metrics. We first change the receptive field from $3\times 3$ patch to $5\times 5$, $7\times 7$, $9\times 9$ and then change the metric method from standard deviation to variation and entropy. The formulation of variation and entropy are listed below, given $3\times 3$ patch as an example.

\begin{equation}
    \begin{gathered} 
        {\rm variance}(\mathbf{c}(u, v))=\frac{1}{9} \sum_{x, y}[\mathbf{c}(x, y)-\Bar{\mathbf{c}}]^2, \\
        {\rm entropy}(\mathbf{c}(u, v))=\sum_{x, y} - \mathbf{c}(x, y) \log\mathbf{c}(x, y), \\
        x\in \{u-1, u, u+1\}, \ y\in \{v-1, v, v+1\}.
    \end{gathered}
\end{equation}

For each combination of patch size and metric method, we visualize the calculated probability distribution of an input image, as shown in Figure~\ref{fig:ablation-patch-visual}. The lighter color represents the larger probability in the grayscale image. The variation of patch sizes and metric methods typically change the difference range of probability distribution between trivial areas and nontrivial areas. Therefore, the hyperparameter of patch size and the different choices of metric methods play the same role as random sampling ratio, which also controls the difference range between small probability and large probability. We also report the PSNR and the training time of each combination on LLFF dataset~\cite{mildenhall2019local} based on Plenoxels~\cite{fridovich2022plenoxels}, as shown in Figure~\ref{fig:ablation-patch-curse}. The results show that the performance of entropy metric is a little worse than standard deviation and variance. However, the performances between different patch sizes do not differ a lot, which means that patch size is an insensitive hyperparameter.

\noindent \textbf{Ablation studies on other modules.} As reported in Table~\ref{tab:ablation-quadtree} and visualized in Figure~\ref{fig:ablation-visual-compare}, we perform extensive ablation studies of various components of our method to demonstrate their effectiveness. Our experiments are based on Plenoxels~\cite{fridovich2022plenoxels} under the Forward-Facing dataset~\cite{mildenhall2019local}. We first remove quadtree subdivision (``w/o quadtree''), which takes more time than Plenoxels to converge.
This is because probability sampling takes some extra time in traversing leaf nodes and calculating probability. We then remove probability distribution sampling (``w/o prob''), which takes less time, but the performance greatly degenerates. Our probability sampling method takes into account the context information of the image and dynamically shoot more rays in more complex regions. Therefore it significantly improves the performance. In the third experiment, we remove all-pixel sampling in the last epoch (``w/o allPixel''), which takes the shortest time and shows the lowest accuracy. It is because that fewer rays are sampled at the last epoch without the all-pixel sampling strategy, which leads to a slight decrease of training time. At last, we remove random sampling ratio (``w/o random''), and we find that both of the time and the performance degenerates slightly. The reason is detailed in the previous ablation section. Our method combines the above strategies and takes a balance between time and accuracy.

%% file: Conclusion.tex
\section{Conclusion}
We introduce a general framework to speed up the learning of radiance fields. Our method successfully leverages a context based probability distribution and adaptive quadtree subdivision to shoot much fewer rays in volume rendering without sacrificing accuracy. Different from the existing approaches which require specific data structures or networks, our method can effectively speed up the training of almost all radiance fields based methods. Our analysis justifies that our method can dynamically shoot more rays to perceive the regions with complex geometry and shoot much fewer rays in simple regions, which significantly reduces redundancy in shooting rays. The evaluation under the widely used benchmarks shows that our method can significantly speed up the learning of radiance fields and achieve comparable accuracy for different methods.